\documentclass[10pt,twocolumn,letterpaper]{article}
\usepackage{subcaption}
\usepackage{iccv}
\usepackage{times}
\usepackage{epsfig}
\usepackage{graphicx}
\usepackage{amsmath}
\usepackage{amssymb}
\usepackage{multirow}
\usepackage{adjustbox}
\usepackage[utf8x]{inputenc} 
\usepackage{csquotes}
\usepackage[table,xcdraw]{xcolor}
\usepackage{nicefrac}       
\usepackage{booktabs}       
\usepackage{amsfonts}       
\usepackage{nicefrac}       
\usepackage{microtype}      
\usepackage{adjustbox}
\usepackage{color}
\usepackage{bm}
\usepackage{multirow}
\usepackage{paralist}

\newcommand*{\affaddr}[1]{\small #1} 
\newcommand*{\affmark}[1][*]{\textsuperscript{#1}}

\usepackage[pagebackref=true,breaklinks=true,colorlinks,bookmarks=false,citecolor=blue,linkcolor=blue,urlcolor=blue]{hyperref}

\iccvfinalcopy 

\def\iccvPaperID{****} 

\def\proposed{\texttt{D2-Net}{}}
\title{D2-Net: Weakly-Supervised Action Localization via Discriminative Embeddings and Denoised Activations}
\author{
Sanath Narayan\affmark[1] \quad
Hisham Cholakkal\affmark[2] \quad 
Munawar Hayat\affmark[3] \quad
Fahad Shahbaz Khan\affmark[2,4] \\
Ming-Hsuan Yang\affmark[5,6,7] \quad
Ling Shao\affmark[1] \vspace{0.01cm}\\
\affaddr{\affmark[1]Inception Institute of Artificial Intelligence} \hspace{0.1cm}
\affaddr{\affmark[2]Mohamed Bin Zayed University of AI} \hspace{0.1cm}
\affaddr{\affmark[3]Monash University} \\
\affaddr{\affmark[4]Linköping University} \hspace{0.1cm}
\affaddr{\affmark[5]University of California, Merced} \hspace{0.1cm}
\affaddr{\affmark[6]Google Research} \hspace{0.1cm}
\affaddr{\affmark[7]Yonsei University}
}

\begin{document}

\maketitle
\begin{abstract}

   This work proposes a weakly-supervised temporal action localization framework, called \proposed, which strives to temporally localize actions using video-level supervision. 
   Our main contribution is the introduction of a novel loss formulation, which jointly enhances the discriminability of latent embeddings and robustness of the output temporal class activations with respect to foreground-background noise caused by weak supervision. 
   The proposed formulation comprises a discriminative and a denoising loss term for enhancing temporal action localization. 
   The discriminative term incorporates a classification loss and utilizes a top-down attention mechanism to enhance the separability of latent foreground-background embeddings. 
   The denoising loss term explicitly addresses the foreground-background noise in class activations by simultaneously maximizing intra-video and inter-video mutual information using a bottom-up attention mechanism. As a result, activations in the foreground regions are emphasized whereas those in the background regions are suppressed, thereby leading to more robust predictions.
   Comprehensive experiments are performed on multiple benchmarks, including THUMOS14 and ActivityNet1.2.
   Our \proposed{} performs favorably in comparison to the existing methods on all datasets, achieving gains as high as 2.3\% in terms of mAP at IoU=0.5 on THUMOS14.
   Source code is available at \url{https://github.com/naraysa/D2-Net}.

\end{abstract}

\section{Introduction\label{sec:introduction}}

Temporal action localization is a challenging problem, which aims to jointly classify and localize the temporal boundaries of actions in videos.~Most existing approaches~\cite{gtad,talnet,rc3d,cdc,ssn,scnn} are based on strong supervision, requiring manually annotated temporal boundaries of actions during training.
In contrast to these strong frame-level supervision based methods, weakly-supervised action localization learns to localize actions in videos, leveraging only video-level supervision.
Weakly-supervised action localization is therefore of greater importance since the manual annotation of temporal boundaries in videos is laborious, expensive and prone to large variations~\cite{action-snippet,action-extent}. 

Existing methods~\cite{hideseek,untrimnets,stpn,wtalc,autoloc} for weakly-supervised action localization typically use video-level annotations in the form of action classes and learn a sequence of class-specific scores, called temporal class activation maps (TCAMs). In general, a classification loss is used to obtain the discriminative foreground regions in TCAMs.
Some approaches~\cite{stpn,wtalc,3cnet,bg-modeling} learn TCAMs using action labels and obtain temporal boundaries via a post-processing step, while others~\cite{autoloc,cleannet} use a TCAM-generating video classification branch along with an explicit localization branch to directly regress action boundaries. 
Nevertheless, the localization performance is heavily dependent on the quality of the TCAMs. 
The quality of TCAMs is likely to improve in fully-supervised settings where frame-level annotations are available. 
Such frame-level information (true foreground and background regions) are unavailable in the weakly-supervised paradigm. In such a paradigm, the predicted foreground regions often overlap with the ground-truth background regions, while predicted background regions are likely to overlap with the ground-truth foreground regions. 
This leads to noisy activations, \ie, false positives and false negatives, in the learned TCAMs. Most existing weakly-supervised action localization methods that learn TCAMs typically rely on separating foreground and background regions (foreground-background separation) and do not explicitly handle its noisy outputs.

\begin{figure}[t]
    \centering
    \includegraphics[width=0.8\columnwidth]{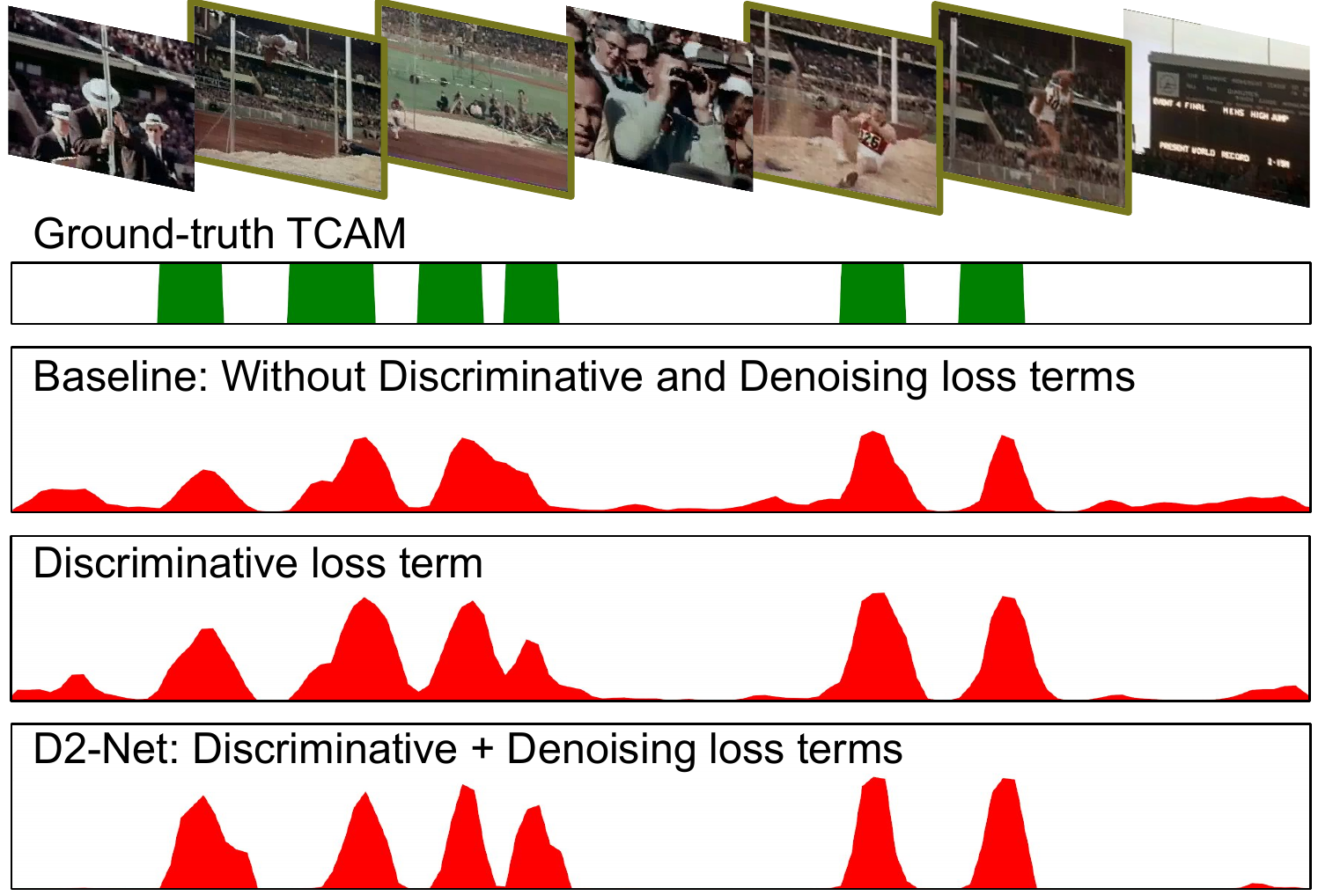}
    \vspace{-0.1cm}
    \caption{\textbf{Impact of our proposed loss formulation} on the quality of the output TCAMs. Compared to the baseline (without our discriminative and denoising loss terms), the introduction of the discriminative loss term improves the separation between foreground and background activations (\eg, third and fourth ground-truth action instance from the left). Furthermore, our final \proposed{} comprising both the discriminative and the denoising loss terms reduces the noise in the TCAMs, leading to more robust TCAMs.  
    \vspace{-0.3cm}}
    \label{fig:denoise_intuition}
\end{figure}

In this work, we address the problem of foreground-background separation along with explicit tackling of noise in TCAMs for weakly-supervised action localization. 
We propose a unified loss formulation that is jointly optimized to classify and temporally localize action snippets (group of frames) in videos. 
Our loss formulation comprises a discriminative and a denoising loss term.
The discriminative loss seeks to maximally separate backgrounds from actions (foregrounds) via interlinked classification and localization learning objectives (Sec.~\ref{sec:integrated_loss}). 
The denoising loss (Sec.~\ref{sec:denoising_loss}) complements the discriminative term by explicitly addressing the foreground-background noise in activations, thereby producing robust TCAMs (see Fig.~\ref{fig:denoise_intuition}).

In our loss formulation, we learn distinct latent embeddings such that their foreground-background separation is maximized based upon the corresponding top-down attention generated from the output TCAMs. 
Furthermore, the embeddings are employed to generate pseudo-labels based on their foreground scores (bottom-up attention). These pseudo-labels are utilized to explicitly handle the noise by emphasizing the corresponding output activations in pseudo-foreground regions, while suppressing the activations in pseudo-background regions.~This pseudo-background suppression and pseudo-foreground enhancement is achieved by maximizing the mutual information (MI) between activations and generated pseudo-labels within an action video (intra-video). Maximizing MI between predicted activations and labels decreases the uncertainty of predictions, leading to more robust predictions. In addition to capturing intra-video MI, our formulation also strives to maximize MI between the action class predictions and video-level ground-truth labels, across videos in a mini-batch (inter-video).

\noindent\textbf{Contributions:} We introduce a weakly-supervised action localization framework, \proposed{}, which incorporates a novel loss formulation that jointly enhances the foreground-background separability and explicitly tackles the noise to robustify the output TCAMs. 
Our main contributions are: 
\begin{compactitem}
\item We introduce a discriminative loss term, which simultaneously aims at video categorization and enhanced foreground-background separation.
\item We introduce a denoising loss term to improve the robustness of TCAMs. Our denoising loss explicitly addresses noise in TCAMs by maximizing the MI between activations and labels within a video (intra-video) \textit{and} across videos (inter-video).  To the best of our knowledge, we are the first to introduce a loss term that simultaneously captures MI across multiple snippets within a video and across all videos in a batch for weakly-supervised action localization.

\item Experiments are performed on multiple benchmarks, including THUMOS14~\cite{thumos14} and ActivityNet1.2~\cite{activitynet}. Our \proposed{} performs favorably against existing weakly-supervised methods on all datasets, achieving gains as high as 2.3\% mAP at IoU=0.5 on THUMOS14. 

\end{compactitem}

\section{Related Work}
Several weak supervision strategies have been explored in the context of action localization, including category labels~\cite{hideseek,stpn,untrimnets,wtalc,autoloc,zhai2020twostream}, sparse temporal points~\cite{point-supervision}, order of actions~\cite{order-constraints-kuehne,order-constraints-bojan}, instance count~\cite{3cnet,star} and single-frame annotations~\cite{ma2020sf}. 
Most existing weakly-supervised action localization methods employ category labels as weak supervision and typically utilize features extracted from backbone networks~\cite{tsn,kinetics} trained on the action recognition task. 
The work of \cite{untrimnets} proposes a selection module for detecting the relevant temporal segments and employs a classification loss for training. 
The Autoloc method~\cite{autoloc} extends~\cite{untrimnets} by adding an explicit localization branch and utilizes an outer-inner contrastive loss for its training.
In contrast,~\cite{wtalc,dml} match similar segments of actions in paired videos by employing classification and similarity-based losses that require multiple videos of same actions in a mini-batch. 
Different from these works, our approach explicitly addresses the issue of large number of easy negatives overwhelming a smaller number of hard positives via sample re-weighting and performs foreground-background separation by inter-linking classification and localization objectives.

\begin{figure*}[t]
    \centering
    \includegraphics[width=0.8\textwidth]{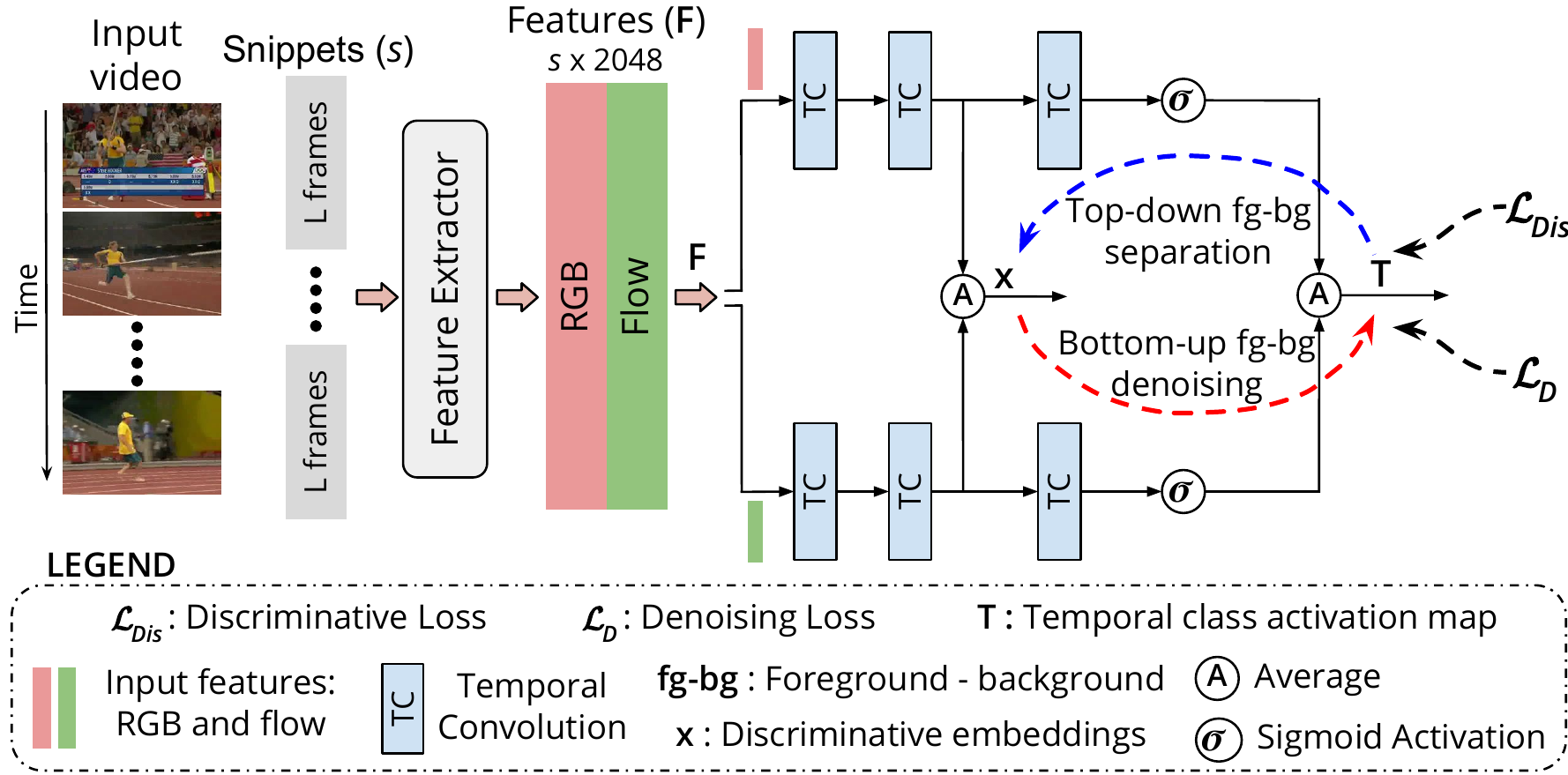}
     \vspace{-1mm}
    \caption{\textbf{Overall architecture} of our \proposed{}. The focus of our design is the introduction of a novel loss formulation that jointly enhances the discriminability of latent embeddings and explicitly addresses the foreground-background noise in the output class activations. 
    The network comprises two identical parallel streams (RGB and flow) consisting of three temporal convolutional \texttt{TC} layers. The second \texttt{TC} layer activations from both streams are averaged to obtain latent embeddings $\mathbf{x}$. The final outputs of both streams are then averaged to obtain the temporal class activation maps (TCAMs) $\mathbf{T}$ of untrimmed input videos. 
    A discriminative loss $\mathcal{L}_{Dis}$ (Sec.~\ref{sec:integrated_loss}) is introduced to enhance the foreground-background separability (\textcolor{blue}{$\dashleftarrow$}) of embeddings $\mathbf{x}$ by utlizing a top-down attention mechanism, 
    in addition to achieving video classification. 
    Furthermore, a denoising loss $\mathcal{L}_D$ (Sec.~\ref{sec:denoising_loss}) is introduced to explicitly address the foreground-background noise (\textcolor{red}{$\dashrightarrow$}) in the class activations of $\mathbf{T}$, by utilizing a bottom-up attention.
    The network is trained jointly using both loss terms $\mathcal{L}_{Dis}$ and $\mathcal{L}_{D}$.\vspace{-0.27cm}
    }
    \label{fig:overall_arch}
\end{figure*}

\noindent\textbf{Snippet-level loss:} While the work of~\cite{bg-modeling} employs a background-aware loss along with a self-guided loss for modeling the background, ~\cite{moniruzzaman2020action} additionally utilizes an iterative multi-pass erasing step for discovering different action segments in TCAMs.
Differently, the training in~\cite{luo2020emmil} alternates between updating a key-instance assignment branch and a classification branch via Expectation Maximization. 
In contrast, the recent work of~\cite{uncertainty2020lee} classifies the foreground/background snippets as in/out-of-distribution based on the feature magnitude and entropy over foreground classes.
However, all these approaches aggregate per-snippet losses for training and do not explicitly capture the mutual information (MI) between the activations and labels, which is likely to be more beneficial due to the absence of snippet-level labels in a weakly-supervised setting.
Different from existing methods~\cite{bg-modeling,moniruzzaman2020action,luo2020emmil,uncertainty2020lee,3cnet,stpn,refineloc,dml}, our approach addresses the problem of foreground-background noise by exploiting both inter- \textit{and} intra-video MI between class activations and corresponding labels, resulting in robust TCAMs. 
To the best of our knowledge, we are the first to propose a weakly-supervised action localization approach that simultaneously captures MI across multiple snippets within a video and across videos in a mini-batch (see also Fig.~\ref{fig:mi_concept}). 

\section{Proposed Method\label{sec:method}}
Our \proposed{} strives to improve the separation of foreground-background feature representations in videos, while jointly enhancing the robustness of output TCAMs \wrt foreground-background noise.
This leads to better differentiation between foreground actions and surrounding background regions, resulting in enhanced action localization in the challenging weakly-supervised setting.
Here, we first present our overall architecture, followed by a detailed description of our proposed losses for training \proposed. 

\noindent\textbf{Overall architecture} of \proposed{} is illustrated in Fig.~\ref{fig:overall_arch}. Given a video $v$, we divide it into non-overlapping snippets of $L=16$ frames each. Features are then extracted to encode appearance (RGB) and motion (optical flow) information. Similar to \cite{stpn,wtalc,3cnet}, we use the Inflated 3D (I3D)~\cite{kinetics} to obtain $d=2048$ dimensional features for each $16$-frame snippet. Let $\mathbf{F} \in \mathbb{R}^{s\times d}$ denote features for a video, where $s$ is the number of snippets. The extracted features become the inputs to our \proposed{}, which comprises two parallel streams for RGB and optical flow. Each stream consists of three temporal convolutional (\texttt{TC}) layers. The first two layers learn latent discriminative embeddings $\mathbf{x}(t) \in \mathbb{R}^{d/2}$ (with time $t \in[1,s]$), from the input features $\mathbf{F}$. The output of the final \texttt{TC} layer is passed through a \textit{sigmoid} activation. Subsequently, the outputs from both streams are averaged to obtain TCAMs $\mathbf{T} \in \mathbb{R}^{s\times C}$ representing a sequence of class-specific scores over time for $C$ action classes. 
The main contribution of our work is the introduction of a novel loss formulation to train the proposed \proposed. 
Our training objective combines a discriminative ($\mathcal{L}_{Dis}$) and a denoising term ($\mathcal{L}_D$), with a balancing weight $\alpha$,
\begin{equation}
    \label{eqn:loss_formulation}
    \mathcal{L} = \mathcal{L}_{Dis} + \alpha\mathcal{L}_D.
\end{equation}
These two loss terms utilize foreground-background attention sequences computed in opposite directions: (\textit{i}) the discriminative loss $\mathcal{L}_{Dis}$ utilizes a top-down attention, which is computed from the output TCAMs (the top-most layer) and (\textit{ii}) the denoising loss $\mathcal{L}_{D}$ utilizes a bottom-up attention, which is derived from the foreground scores of the latent embeddings (intermediate layer features). 
We describe these losses in detail in Sec.~\ref{sec:integrated_loss} and~\ref{sec:denoising_loss}.

\subsection{\hspace{-0.07em}Foreground-Background Discriminability:$\mathcal{L}_{Dis}$\label{sec:integrated_loss}}
In this work, we introduce a discriminative loss ($\mathcal{L}_{Dis}$) to learn separable class-agnostic foreground and action-free background feature representations, in terms of latent embeddings, using a top-down attention from the TCAMs. 
The embedding of a video with $s$ snippets is defined by a weighted temporal pooling based on the class activations $\mathbf{T} \in \mathbb{R}^{s \times C}$.
Let the top-down foreground attention $\bm{\lambda}(t)= \max_{c}\mathbf{T}[t,c]$ denote the maximum foreground activation across all action classes $c\in\{1,\ldots,C\}$, where $t \in [1,s]$ and $C$ is the number of classes. Then, the class-agnostic foreground and background embeddings are:
\begin{equation}
    \mathbf{x}_{fg} = \sum\limits_{\bm{\lambda}(t)>\tau} \bm{\lambda}(t)\mathbf{x}(t) , \quad
    \mathbf{x}_{bg} = \sum\limits_{\bm{\lambda}^b(t)>\tau} \bm{\lambda}^b(t)\mathbf{x}(t),
    \label{eqn:feature_aggregate}
\end{equation}
where $\tau{=}0.5$ and $\bm{\lambda}^b(t){=}1{-}\bm{\lambda}(t)$ is the background attention.
Maximizing the distance between foreground and background embeddings enhances the separability of the corresponding output activations, leading to improved localization.
In addition, different sets of action classes are likely to share certain characteristics among them \eg, \textit{Hammer Throw} and \textit{Discus Throw} have similar spatial context and motion. Hence, clustering foreground embeddings amongst themselves at a coarse level is likely to aid ``coarse-to-fine" snippet-level classification.
Similarly, clustering background embeddings helps in learning an approximate universal background embedding, which is likely to aid in generalization at test time to new backgrounds.
Hence, three weight terms, $w_{fb}, w_{fg}$ and $w_{bg}$, are introduced in our $\mathcal{L}_{Dis}$, targeting foreground-background separation, foreground grouping and background grouping, respectively. They are defined as: 
\begin{align}
w_{fb} & = \max(0,\cos (\mathbf{x}_{fg},\tilde{\mathbf{x}}_{bg})), \notag \\
w_{fg} & = \gamma (1-\cos (\mathbf{x}_{fg},\tilde{\mathbf{x}}_{fg})), \notag \\
w_{bg} & = \gamma (1-\cos (\mathbf{x}_{bg},\tilde{\mathbf{x}}_{bg})), 
\label{eqn:cosine_wts}
\end{align}
%
%
%
where $\mathbf{x}$ and $\tilde{\mathbf{x}}$ denote embeddings from different videos in a mini-batch. Here, $\gamma$ denotes the intra-class compactness weight used for grouping same class (foreground \vs background) embeddings.
Alongside robust localization, our other objective is the multi-label classification of action categories. 
A major challenge is introduced by the class-imbalance problem, where easy background snippets overwhelmingly outnumber the hard foregrounds. 
To address this, inspired by the focal loss for object detection~\cite{focal_loss}, we propose to include penalty terms based on the weights (Eq.~\ref{eqn:cosine_wts}), in our $\mathcal{L}_{Dis}$.
To this end, a video-level prediction $\mathbf{p} \in \mathbb{R}^C$ is obtained by performing a temporal \emph{top-k} pooling on $\mathbf{T}$. Our $\mathcal{L}_{Dis}$ term, which jointly addresses the class-imbalance and enhances foreground-background separation, is defined by
\begin{align}
    \label{eqn:discriminative_loss}
    \mathcal{L}_{Dis} & = - {\sum_{c:\mathbf{y}[c]=1}}(1-\mathbf{p}[c] +w_{fg}+w_{fb})^\beta\log(\mathbf{p}[c]) \notag \\
    & - {\sum_{c:\mathbf{y}[c]=0}}(\mathbf{p}[c]+w_{bg}+w_{fb})^\beta\log(1-\mathbf{p}[c]),
\end{align}
where $\mathbf{y} \in \{0,1\}^C$ denotes the video-level label and $\beta$ is the focusing parameter. The first term in Eq.~\ref{eqn:discriminative_loss} denotes the loss for a positive action class, while the second term incorporates the loss for a negative class. The weight term $w_{fb}$ (see Eq.~\ref{eqn:cosine_wts}) is added for both positive action classes and background classes since it represents the foreground-background separation. 
The terms $w_{fg}$ and $w_{bg}$ enhance intra-class compactness for the positive and background classes, respectively. 
The first term in Eq.~\ref{eqn:discriminative_loss} indicates that the loss due to a positive action class $c$ is low only when (i) its predicted probability $\mathbf{p}[c]$ is high, and (ii) the foreground grouping $w_{fg}$ and foreground-background separation $w_{fb}$ for the corresponding video are both simultaneously low. A similar observation holds in the second term for the negative class. Thus, $\mathcal{L}_{Dis}$ enhances the discriminability of embeddings $\mathbf{x}(t)$ by encouraging foreground-background separation while simultaneously achieving classification.

\subsection{Robust Temporal Class Activation Maps: $\mathcal{L}_D$\label{sec:denoising_loss}}
Our discriminative loss $\mathcal{L}_{Dis}$ improves action localization by enhancing the distinctiveness of latent embeddings.
However, the temporal locations of true foreground regions are unknown under weak supervision, resulting in noisy output temporal class activations (and noisy top-down attention) learned from video-level labels.
Consequently, the foreground and background embeddings ($\mathbf{x}_{fg}$ and $\mathbf{x}_{bg}$), learned from the top-down attention $\bm{\lambda}(t)$, are likely to be noisy. 
Our goal is to explicitly reduce this foreground-background noise caused by the absence of snippet-level labels and improve the robustness of the output class activations. 
To this end, we introduce a denoising loss $\mathcal{L}_D$ comprising a novel pseudo-Determinant based Mutual Information (pDMI) loss. Our $\mathcal{L}_D$ exploits both intra- and inter-video mutual information (MI) between the class activations and corresponding labels.

Our pseudo-Determinant based Mutual Information (pDMI) loss is inspired by the Determinant based Mutual Information (DMI)~\cite{dmi_neurips19}. The original DMI, proposed for multi-class classification, is computed as the determinant of a joint distribution matrix, \ie, $\text{DMI}(\mathbf{P},\mathbf{Y}){=}|\det(\mathbf{U})|$. Here, $\mathbf{U}=\nicefrac{1}{n}\mathbf{PY}$ is the joint distribution over the predicted posterior probabilities $\mathbf{P}$ and the ground-truth (noisy) labels $\mathbf{Y}$. The matrices $\mathbf{P}$ and $\mathbf{Y}$ are of sizes $C\times n$ and $n \times C$, where $n$ denotes the mini-batch size and $C$ the number of classes. The DMI loss $\mathcal{L}_{dmi}$ is defined as
\begin{equation}
    \label{eqn:original_dmi_loss}
    \mathcal{L}_{dmi} = -\mathbb{E}[\log(|\det(\mathbf{U})|)],
\end{equation}
where $\mathbb{E}$ denotes Expectation.
Note that $\mathcal{L}_{dmi}$ depends on the determinant of $\mathbf{U}$. To ensure a non-zero $\det(\mathbf{U})$, the label matrix $\mathbf{Y}$ must be full-rank, \ie, a mini-batch must contain instances from all classes. This is prohibitive for a large number of classes. Such a mini-batch sampling for action localization also leads to memory issues in GPUs due to the long duration of untrimmed videos in the dataset, especially when capturing inter-video MI.

\begin{figure}[t]
    \centering
    \includegraphics[clip=true, trim=0em 0em 0em 4.2em,width=0.47\columnwidth]{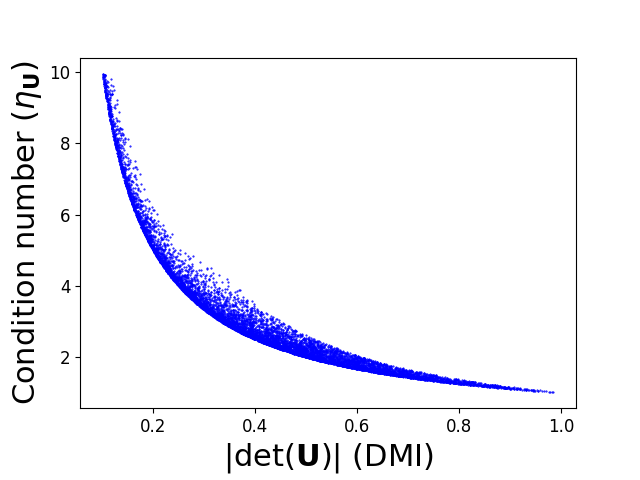}
    \includegraphics[clip=true, trim=0em 0em 0em 4.2em,width=0.47\columnwidth]{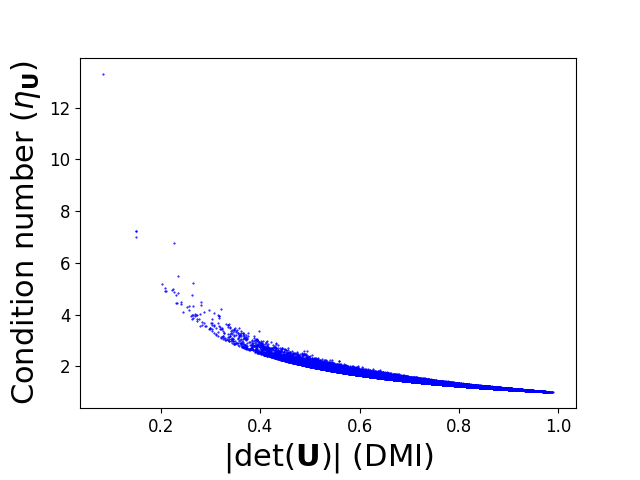}
    \vspace{-0.12cm}
    \caption{\textbf{Condition number ($\eta_{\mathbf{U}}$) \vs Determinant ($|\text{det}(\mathbf{U})|$)} for joint distribution matrices $\mathbf{U}$. On the left: $25$k randomly sampled $\mathbf{U}$. On the right: $\mathbf{U}$ obtained during our snippet-level training. In both cases, minimizing $\eta_{\mathbf{U}}$ leads to maximizing $|\text{det}(\mathbf{U})|$ (DMI).
    \vspace{-0.3cm}}
    \label{fig:eta_det}
\end{figure}

Our pDMI loss overcomes these limitations and ensures a non-degenerate value of DMI by avoiding an explicit computation of the determinant.
To this end, we observe that for the DMI loss to tend to zero, the determinant of the joint distribution $|\det(\mathbf{U})|$ must tend to one.
Formally,
\begin{equation}
    \mathcal{L}_{dmi} \xrightarrow{} 0 \implies |\det(\mathbf{U})| \xrightarrow{} 1 \implies \mathbf{U} \xrightarrow{} \mathbf{I}. 
\end{equation}
As a result, DMI is maximum when $|\det(\mathbf{U})|{=}1$, with the identity matrix $\mathbf{I}$ as an optima for $\mathbf{U}$ of size $C\times C$ (since elements of $\mathbf{U} {\in} [0,1]$). Furthermore, the condition number $\eta$ for the optimal solution $\mathbf{I}$ is minimum, \ie, $\eta{=}1$. Hence, instead of maximizing  $|\det(\mathbf{U})|$, we can alternatively minimize its $\eta$. In effect, $\mathbf{U}$ becomes better-conditioned and this improves the robustness of the activations towards label noise. 
The proposed pDMI loss $\mathcal{L}_{pdmi}$ is then given by
\begin{equation}
    \label{eqn:pdmi_loss}
    \mathcal{L}_{pdmi} = \mathbb{E}[\log(\text{pDMI}(\mathbf{P},\mathbf{Y}))] = 
    \mathbb{E}[\log(\eta_\mathbf{U})],
\end{equation}
where $\eta_\mathbf{U}$ denotes the condition number of $\mathbf{U}$. Since the rank of $\mathbf{U}$ is $r\leq C$, $\eta_\mathbf{U}$ is computed as $\sigma_1/\sigma_r$, where $\{\sigma_1, \ldots, \sigma_r\}$ are non-zero singular values of $\mathbf{U}$. Thus, our pDMI loss avoids an explicit computation of the determinant and overcomes the limitations of the standard DMI. 
Fig.~\ref{fig:eta_det} shows plots of $\eta_{\mathbf{U}}$ \vs $|\text{det}(\mathbf{U})|$ for joint distribution matrices $\mathbf{U}$ that are randomly sampled (left) and encountered during intra-video MI training (right, described in Sec.~\ref{sec:intra_inter_MI}).
It can be observed that minimizing $\eta_\mathbf{U}$ indeed maximizes $|\text{det}(\mathbf{U})|$, \ie, DMI, in turn maximizing MI.
Consequently, our pDMI serves as a promising alternative to the original DMI when optimizing with noisy temporal action labels.

\begin{figure}[t]
    \centering
    \includegraphics[clip=true, trim=0em 0em 0em 0em, width=1\columnwidth, keepaspectratio]{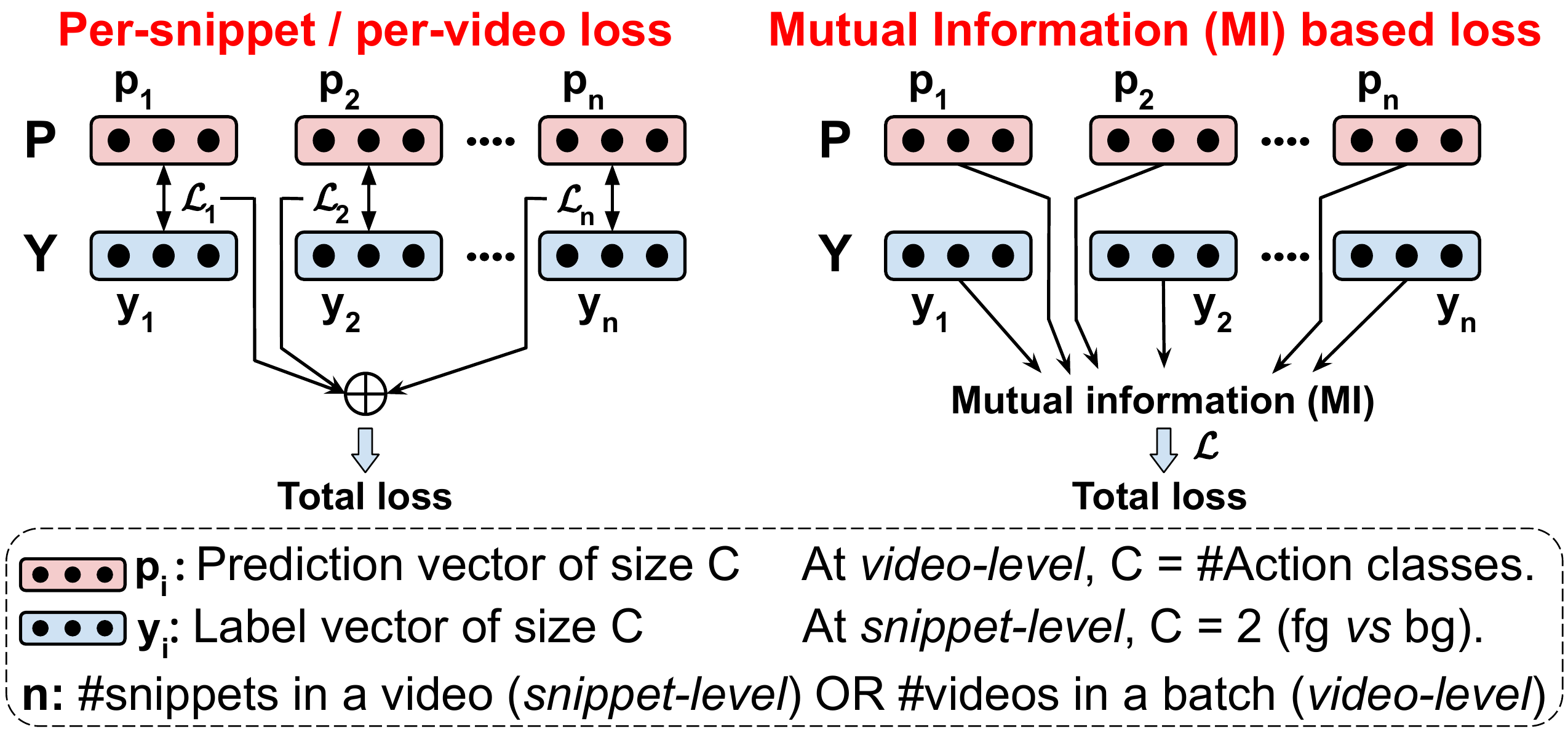}\vspace{-0.1cm}
    \caption{\label{fig:mi_concept}\textbf{A conceptual illustration of loss computation} with (on the right) and without (on the left) capturing mutual information (MI). Typically, existing methods compute the loss without MI (\eg, cross-entropy loss) by aggregating individual losses ($\mathcal{L}_i$) between prediction $\mathbf{p}_i$ and labels $\mathbf{y}_i$ either at a per-video or per-snippet level. Instead, we compute a collective loss across (i) all snippets within a video (snippet-level) \textit{and} (ii) all videos in a batch (video-level), by capturing the MI between predictions ($\mathbf{P}$) and labels ($\mathbf{Y}$).\vspace{-0.25cm}}
\end{figure}

\subsubsection{Snippet-level and Video-level Noise Removal\label{sec:intra_inter_MI}}
To robustify the TCAMs, we employ our $\mathcal{L}_{pdmi}$ at two levels: (\textit{i}) snippet-level to exploit intra-video MI, and (\textit{ii}) video-level to exploit inter-video MI. 
Snippet-level denoising incorporates a bottom-up attention to emphasize the foreground activations, while suppressing the background ones by capturing the MI between the temporal activations and corresponding foreground labels within a video. 
On the other hand, the video-level denoising step exploits MI between the video representations and corresponding labels, across videos, to achieve the same objective. 
Fig.~\ref{fig:mi_concept} shows a conceptual illustration of loss computation with and without capturing MI. \\
\noindent\textbf{Snippet-level joint distribution:}
It captures the MI between the foreground-background activations and the snippet-level pseudo-labels within a video. For this, we utilize a bottom-up attention mechanism, which encodes the foreground scores $\bm{\lambda}'(t)$ of latent embeddings $\mathbf{x}(t)$ for the corresponding snippets. The scores $\bm{\lambda}'(t)$ are computed \wrt a reference background embedding $\mathbf{x}_{ref}$ and are given by
\begin{equation}
    \label{eqn:bottom_up_attention}
    \bm{\lambda}'(t) = 0.5(1 - \cos (\mathbf{x}(t), \mathbf{x}_{ref})), \quad t \in [1, s],
\end{equation}
where $\mathbf{x}_{ref}^{[m]}\!=\!0.9\mathbf{x}_{ref}^{[m-1]} {+} 0.1\mathbf{x}_{bg}^{\mu,[m]}$ is progressively computed as a running mean of $\mathbf{x}_{bg}$ over $m$ iterations. Here, $\mathbf{x}_{bg}^{\mu,[m]}$ denotes the mean of the background embeddings in a mini-batch at iteration $m$. 
Let $t_f {=} \{t{:}\bm{\lambda}^{'}(t) {>} 0.5\}$ and $t_b {=} \{t{:}\bm{\lambda}^{'}(t) {<} 0.5\}$ denote the time instants for selecting the foreground and background activations \wrt $\bm{\lambda}^{'}(t)$. Using the pseudo-foreground temporal locations $t_f$, a row matrix
$\bm{\lambda}_f$ of width $n_f {=} |t_f|$ is constructed using top-down attention  $\bm{\lambda}(t), t {\in} t_f$. Similarly, $\bm{\lambda}_b$ of width $n_b {=} |t_b|$ is constructed for the pseudo-background snippets. 
Then, the prediction matrix $\mathbf{P}_1$ and pseudo-label matrix $\mathbf{Y}_1$ are given by
\begin{equation}
    \label{eqn:snippet_joint}
    \mathbf{P}_1 = \begin{bmatrix} \bm{\lambda}_f  &  \bm{\lambda}_b \\  1 - \bm{\lambda}_f & 1 -\bm{\lambda}_b \end{bmatrix} , \quad
    \mathbf{Y}_1 = \nicefrac{1}{z} \begin{bmatrix} \bm{1}_{n_f} & \bm{0}_{n_f} \\ \bm{0}_{n_b} & \bm{1}_{n_b} \end{bmatrix}, 
\end{equation}
where $z{=}n_f{+}n_b$, $\mathbf{P}_1 {\in} \mathbb{R}^{2\times z} $, $\mathbf{Y}_1 {\in} \mathbb{R}^{z \times 2}$, $\bm{1}_k$ and $\bm{0}_k$ are $k$ dimensional column vectors of ones and zeros.
The snippet-level joint distribution is then defined as $\mathbf{U}_1= \mathbf{P}_1\mathbf{Y}_1$. 
\\
\textbf{Video-level joint distribution:}
Here, the noise stems from the video-level prediction $\mathbf{p} \in \mathbb{R}^C$ and is predominantly caused by the temporal \textit{top-k} pooling. Under the weakly-supervised setting, all the \textit{top-k} locations predicted for an action class need not necessarily belong to that class. Moreover, actions in untrimmed videos may not span $k{=}\lceil s/8\rceil$ snippets.
Hence, denoising the video-level prediction $\mathbf{p}$ eventually robustifies the output class activations at the snippet-level. 
Let the prediction $\mathbf{P}_2$ and label $\mathbf{Y}_2$ be 
\begin{equation}
    \label{eqn:video_joint}
    \mathbf{P}_2 = \begin{bmatrix} \mathbf{p}_1, \ldots, \mathbf{p}_n \end{bmatrix} \hspace{0.15cm} \text{and} \hspace{0.15cm}
    \mathbf{Y}_2 = \nicefrac{1}{n}\begin{bmatrix} \mathbf{y}_1, \ldots, \mathbf{y}_n \end{bmatrix}^{\top}, 
\end{equation}
where $\mathbf{p}_i \in \mathbb{R}^C$ and $\mathbf{y}_i \in \{0,1\}^C$ denote the video-level prediction and associated label of $i$-th video in a mini-batch. Then, the video-level joint distribution that captures the MI between class activations and action classes across videos is  $\mathbf{U}_2{=}\mathbf{P}_2\mathbf{Y}_2$. We finally define our denoising loss as
\begin{align}
    \mathcal{L}_D & = \mathcal{L}_{DS} + \mathcal{L}_{DV}  \\ & = \mathbb{E}[\log(\text{pDMI}(\mathbf{P}_1,\mathbf{Y}_1))]   + \mathbb{E}[\log(\text{pDMI}(\mathbf{P}_2,\mathbf{Y}_2))], \notag
\end{align}
where the pDMI loss is given by Eq.~\ref{eqn:pdmi_loss}. Here, $\mathcal{L}_{DS}$ and $\mathcal{L}_{DV}$ denote the snippet-level and video-level losses.~Thus, our denoising loss improves the TCAMs, at the snippet-level and video-level, by making them robust to the foreground-background noise under the weakly-supervised setting.

\subsection{Inference: Action Localization from TCAMs}
At inference, given a video, \proposed{} outputs a bottom-up attention sequence $\bm{\lambda}^{'}$ (Eq.~\ref{eqn:bottom_up_attention}) of length $s$ and a class activation map $\mathbf{T}$ of size $s \times C$. We perform \textit{top-k} pooling to obtain the predicted class probabilities $\mathbf{p} \in \mathbb{R}^C$, which are then used to find the relevant action classes above a threshold $p_{th}=0.5\max(\mathbf{p})$. For every relevant class $c$, its corresponding class activations $\mathbf{T}_c\in \mathbb{R}^s$ are multiplied element-wise with $\bm{\lambda}^{'}\in \mathbb{R}^s$ to obtain a refined sequence $\mathbf{r}_c=\bm{\lambda}^{'}\mathbf{T}_c$. The snippets with activations above a threshold are retained and a 1-D connected component is used to obtain segment proposals. 
Multiple thresholds are used to obtain a larger pool of proposals. Each proposal is then scored using the contrast between the mean activation of the proposal itself and its surrounding areas~\cite{autoloc}, $S = S_{i} - S_{o}$, where $S_i$ and $S_o$ respectively denote the mean activation of the proposal and its neighboring background. The neighboring background is obtained by inflating the proposal on either side by $25\%$ of its width, as in~\cite{autoloc}.
Proposals with high overlap are removed using class-wise NMS. 
Only high-scoring proposals (\ie, $S>S_{th}$) are retained as final detections.

\begin{table}[t]
\centering
\caption{\label{tab:sota_th14}\textbf{State-of-the-art comparison} on the THUMOS14 dataset. Methods with superscript `+' require strong frame-level supervision for training. Our \proposed{} performs favorably in comparison to existing weakly-supervised methods and achieves consistent improvements, in terms of mean average precision (mAP).}
\setlength{\tabcolsep}{10pt}
\adjustbox{width=\columnwidth}{
\begin{tabular}{lccccc}
\toprule
\multicolumn{1}{l}{\multirow{2}{*}{\textbf{Approach}}} & \multicolumn{5}{c}{\textbf{mAP @ IoU}} \\ 
\multicolumn{1}{c}{} & \textbf{0.1} & \textbf{0.2} & \textbf{0.3} & \textbf{0.4} & \textbf{0.5}  \\ 
 \midrule
\texttt{R-C3D}~\cite{rc3d}${}^{+}$  & 54.5 & 51.5 & 44.8 & 35.6 & 28.9 \\
\texttt{GTAD}~\cite{gtad}${}^{+}$  & - & - & 54.5 & 47.6 & 40.2 \\
\texttt{TAL-Net}~\cite{talnet}${}^{+}$  & 59.8 & 57.1 & 53.2 & 48.5 & 42.8 \\ 
\texttt{P-GCN}~\cite{zeng2019graph}${}^{+}$  & \textbf{69.5} & \textbf{67.8} & \textbf{63.6} & \textbf{57.8} & \textbf{49.1} \\ 
\midrule
\texttt{Autoloc}~\cite{autoloc}  & - & - & 35.8 & 29.0 & 21.2 \\ 
\texttt{W-TALC}~\cite{wtalc} & 53.7 & 48.5 & 39.2 & 29.9 & 22.0 \\
\texttt{CMCS}~\cite{cmcs-modeling}  & 57.4 & 50.8 & 41.2 & 32.1 & 23.1 \\ 
\texttt{BM}~\cite{bg-modeling}  & 64.2 & 59.5 & 49.1 & 38.4 & 27.5 \\
\texttt{3C-Net}~\cite{3cnet}  & 59.1 & 53.5 & 44.2 & 34.1 & 26.6 \\
\texttt{BaS-Net}~\cite{basnet} & 58.2 & 52.3 & 44.6 & 36.0 & 27.0 \\
\texttt{DGAM}~\cite{dgam} & 60.0 & 54.2 & 46.8 & 38.2 & 28.8  \\
\texttt{DML}~\cite{dml} & 62.3 & - & 46.8 & - & 29.6 \\
\texttt{A2CL-PT}~\cite{min2020adversarial} & 61.2 & 56.1 & 48.1 & 39.0 & 30.1 \\
\texttt{EM-MIL}~\cite{luo2020emmil} & 59.1 & 52.7 & 45.5 & 36.8 & 30.5 \\
\texttt{ACM-BANet}~\cite{moniruzzaman2020action} & 64.6 & 57.7 & 48.9 & 40.9 & 32.3 \\
\texttt{HAM-Net}~\cite{islam2021hybrid} & 65.4 & 59.0 & 50.3 & 41.1 & 31.0 \\
\texttt{UM}~\cite{uncertainty2020lee} & \textbf{67.5} & \textbf{61.2} & \textbf{52.3} & \textbf{43.4} & 33.7 \\
\texttt{ASL}~\cite{ma2021weakly} & 67.0 & - & 51.8 & - & 31.1 \\
\texttt{CoLA}~\cite{zhang2021cola} & 66.2 & 59.5 & 51.5 & 41.9 & 32.2 \\
\textbf{Ours: \proposed} & 65.7 & 60.2 & \textbf{52.3} & \textbf{43.4} & \textbf{36.0} \\ \bottomrule
\end{tabular}
}
\vspace{-0.2cm}
\end{table}

\section{Experiments}
\noindent\textbf{Datasets}: We evaluate \proposed{} on multiple challenging temporal action localization benchmarks.
The \textbf{THUMOS14}~\cite{thumos14} dataset contains temporal annotations for $200$ validation and $212$ test videos from $20$ action categories. The dataset is challenging since each video contains $15$ action instances on an average. As in~\cite{wtalc,refineloc}, the validation and test set are used for training and evaluating, respectively. 
The \textbf{ActivityNet1.2}~\cite{activitynet} dataset has annotations of $100$ categories in $4819$ training and $2383$ validation videos, with $1.5$ activity instances per video on an average. As in~\cite{autoloc,wtalc}, we use the training and validation sets to respectively train and evaluate.\\
\textbf{Implementation details}: 
For each snippet, $2048$-$d$ features are extracted from RGB and Flow I3D models pre-trained on Kinetics~\cite{kinetics}. 
The kernel size and dilation rate of the temporal convolutional layers are: ($3$, $1$) for THUMOS14 and ($5$, $2$) for ActivityNet1.2. The first two convolutions in each stream are followed by a leaky ReLU with $0.2$ negative slope. Our \proposed{} is trained with a mini-batch size of $10$ for $20$K iterations, using the Adam~\cite{adam} optimizer with a $10^{-4}$ learning rate and $0.005$ weight decay.
The $k$ for $top$-$k$ is set to $\lceil s/8\rceil$, as in~\cite{wtalc,3cnet}.
All the hyperparameters are chosen via cross-validation. The balancing parameter $\alpha$ is set to $0.2$ and $10^{-3}$ for THUMOS14 and ActivityNet1.2. The intra-class compactness weight $\gamma$ and focusing parameter $\beta$ are set to $0.01$ and $2$ for both datasets. Multiple thresholds from $0.025$ to $0.5$ with increments of $0.025$ are used for proposal generation. The NMS threshold is set to $0.5$ while the score threshold $S_{th}$ for retaining detections in a video is set to $10\%$ of the maximum proposal score in that video.

\begin{table}[t]
\centering
\caption{\label{tab:sota_acn}\textbf{State-of-the-art comparison} on the ActivityNet1.2 dataset. Our \proposed{} performs favorably compared to existing weakly-supervised approaches. Furthermore, our \proposed{} performs comparably to \texttt{SSN}~\cite{ssn}, which is trained with strong supervision (denoted with superscript `+'). AVG denotes the mean of the mAP values for IoU in $[0.5, 0.95]$ with steps of $0.05$.}
\setlength{\tabcolsep}{14pt}
\adjustbox{width=\columnwidth}{
\begin{tabular}{lccc|c}
\toprule
\multicolumn{1}{l}{\multirow{2}{*}{\textbf{Approach}}} & \multicolumn{3}{c}{\textbf{mAP @ IoU}} & \multicolumn{1}{|c}{\multirow{2}{*}{\textbf{AVG}}} \\ 
\multicolumn{1}{c}{} & \textbf{0.5} & \textbf{0.75} & \textbf{0.95} &    \\ 
 \midrule
\texttt{SSN}~\cite{ssn}${}^{+}$ & \textbf{41.3} & \textbf{27.0} & \textbf{6.1} & \textbf{26.6} \\ \midrule
\texttt{DML}~\cite{dml} & 35.2 & - & - & -  \\
\texttt{EM-MIL}~\cite{luo2020emmil} & 37.4 & - & - & 20.3  \\
\texttt{CMCS}~\cite{cmcs-modeling} & 36.8 & 22.0 & 5.6 & 22.4  \\
\texttt{3C-Net}~\cite{3cnet} & 37.2 & - & - & 21.7  \\
\texttt{BaS-Net}~\cite{basnet} & 38.5 & 24.2 & 5.6 & 24.3 \\
\texttt{DGAM}~\cite{dgam} & 41.0 & 23.5 & 5.3 & 24.4 \\
\texttt{UM}~\cite{uncertainty2020lee} & 41.2 & \textbf{25.6} & \textbf{6.0} & 25.9 \\
\texttt{ASL}~\cite{ma2021weakly} & 40.2 & - & - & 25.8 \\
 \textbf{Ours: \proposed} &\textbf{42.3} & 25.5 & 5.8 & \textbf{26.0} \\ \bottomrule
\end{tabular}
}
\vspace{-0.2cm}
\end{table}

\subsection{State-of-the-art Comparison \label{sec:sota_compare}}
Tab.~\ref{tab:sota_th14} and \ref{tab:sota_acn} compare \proposed{} with state-of-the-art methods on THUMOS14 and ActivityNet1.2, respectively. Methods with '$+$' require strong supervision for training.\\
\noindent\textbf{THUMOS14:} 
Similar to ours, all weakly-supervised methods in Tab.~\ref{tab:sota_th14} use an I3D backbone, except \texttt{Autoloc}~\cite{autoloc}, which uses TSN~\cite{tsn}. While \texttt{BM}~\cite{bg-modeling}  considers an additional background class, \texttt{DGAM}~\cite{dgam} extends BM using a VAE~\cite{kingma13iclr}. Although \texttt{DML}~\cite{dml} and \texttt{EM-MIL}~\cite{luo2020emmil} achieve a promising mAP of $29.6$ and $30.5$ at IoU=0.5, they do not generalize well to ActivityNet1.2 (see Tab.~\ref{tab:sota_acn}). As discussed earlier, the recent work of \texttt{UM}~\cite{uncertainty2020lee} employs out-of-distribution detection of background snippets. We also empirically validate the complementarity of our approach with \texttt{UM} by intergrating the loss terms and observe an average gain of $1\%$ mAP across different IoUs. Our \proposed{} performs well against existing weakly-supervised approaches, including the recent \texttt{CoLA}~\cite{zhang2021cola} and \texttt{ASL}~\cite{ma2021weakly}. Our approach achieves an absolute gain of $2.3\%$ at IoU=$0.5$ over the best existing method (\texttt{UM}). Moreover, promising localization performance is obtained at other IoU thresholds.\\
\noindent\textbf{ActivityNet1.2:} 
Similar to our \proposed, all weakly-supervised methods in Tab.~\ref{tab:sota_acn} use I3D backbone. Following standard evaluation protocol~\cite{activitynet}, we report the mean of the mAP scores (denoted as \texttt{AVG}) at different IoU thresholds ($[0.5,0.95]$ in steps of $0.05$). The generative modeling based approach \texttt{DGAM}~\cite{dgam} and background suppression based \texttt{BaS-Net}~\cite{basnet} perform comparably, achieving mean mAP scores of $24.4$ and $24.3$, respectively. In comparison, the recent approaches such as UM~\cite{uncertainty2020lee} and ASL~\cite{ma2021weakly} achieve localization performances of $25.9$ and $25.8$, respectively, in terms of mean mAP. Our proposed \proposed{} performs comparably against these existing approaches and achieves a promising localization performance of $26.0$ mean mAP. Additional results are provided in the appendix.\\
%
%
%

\begin{figure*}[t]
    \centering
    \includegraphics[width=0.94\textwidth,height=0.194\textwidth]{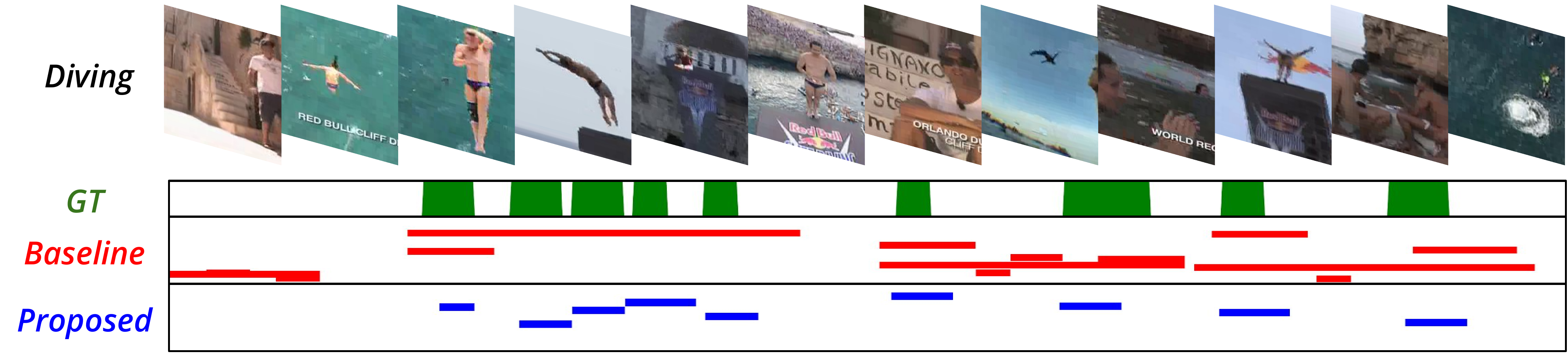}
    \includegraphics[width=0.94\textwidth,height=0.194\textwidth]{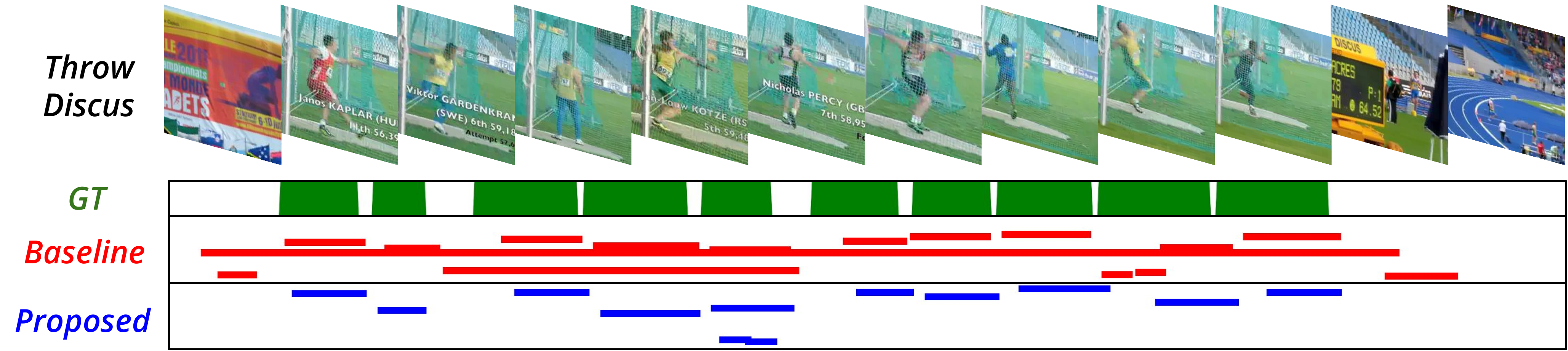}
    \includegraphics[width=0.94\textwidth,height=0.194\textwidth]{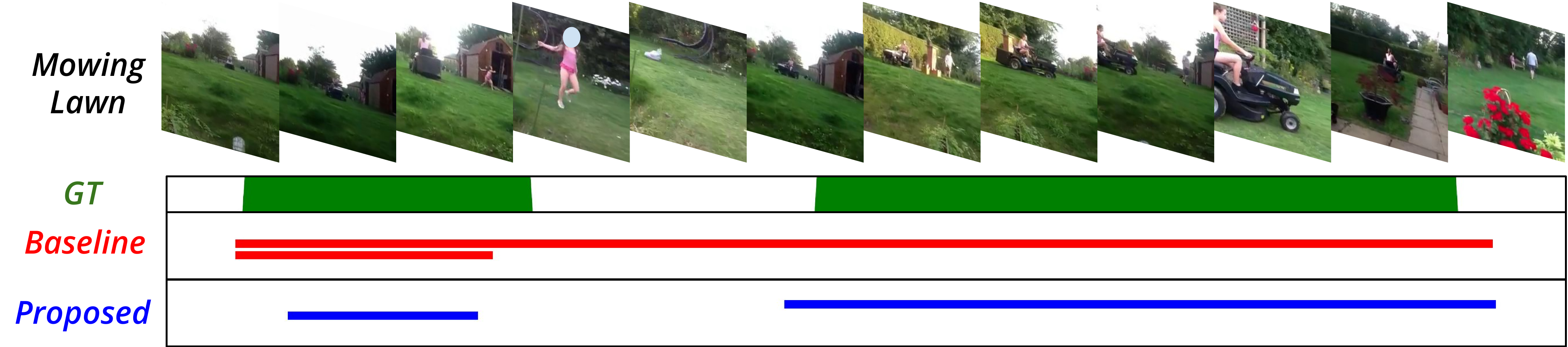}
    \caption{\textbf{Qualitative temporal action localization results} of our proposed \proposed{} on example test videos, with \textit{Diving}, \textit{Throw Discus} actions from THUMOS14, and \textit{Mowing Lawn} activity from ActivityNet1.2. For each video, example frames (top row), ground-truth GT segments (green), baseline detections (red) and \proposed{} detections (blue) are shown. The height of a detection is indicative of its score. The \texttt{Baseline} incorrectly merges multiple GT instances, has false positives in background regions and falsely detects the presence of the activity over the entire video length. Our \proposed{} correctly detects multiple instances (\eg, $1$ to $5$ GT in \textit{Diving}, $3$ to $5$ in \textit{Throw Discus}) and suppresses most false positives in the background regions, achieving promising localization performance.
    \vspace{-0.2cm}}
    \label{fig:qual_res_acn}
\end{figure*}
%

%
%
%
%
\begin{table}[t]
\centering
\caption{\label{tab:ablation}\textbf{Performance comparison} by replacing our two loss terms ($\mathcal{L}_{Dis}$ and $\mathcal{L}_{D}$) in the proposed \proposed{} with either the standard cross-entropy loss ($\mathcal{L}_{CE}$) or the focal loss ($\mathcal{L}_F$). In addition, we also show the performance of our \proposed{} with only $\mathcal{L}_{Dis}$.
Results are shown in terms of mAP and F1 score at IoU=$0.5$, on THUMOS14. Replacing the proposed loss terms in our framework with $\mathcal{L}_{CE}$ and $\mathcal{L}_F$ results in mAP scores at IoU=$0.5$ of 23.0 and 26.7, respectively. 
Our \proposed{} with the discriminative loss term $\mathcal{L}_{Dis}$ achieves consistent improvement in performance over $\mathcal{L}_F$ with an absolute gain of 5.5\% in terms of mAP at IoU=$0.5$. Furthermore, our final \proposed{} comprising both loss terms ($\mathcal{L}_{Dis}$ and $\mathcal{L}_{D}$) achieves the best performance with absolute gains of 12.9\% and 9.2\% in terms of mAP at IoU=$0.5$ over $\mathcal{L}_{CE}$ and $\mathcal{L}_F$, respectively.
}
\adjustbox{width=\columnwidth}{
\begin{tabular}{lccccccc}
\toprule
\multirow{2}{*}{\textbf{Loss term}} &  \multicolumn{5}{c}{\textbf{mAP @ IoU}} & \textbf{F1} \\ 
 &  \textbf{0.1} & \textbf{0.2} & \textbf{0.3} & \textbf{0.4} & \textbf{0.5}  & \textbf{} \\ \midrule
 $\mathcal{L}_{CE}$ & 55.0 & 47.6 & 38.7 & 30.7 & 23.0 & 23.5 \\
 $\mathcal{L}_{F}$ &  58.8 & 52.4 & 44.3 & 35.7 & 26.7 & 27.2 \\ \midrule
 $\mathcal{L}_{Dis}$ &  65.4 & 59.7 & 50.1 & 40.4 & 32.2 & 30.7  \\
 \textbf{\proposed:} $\mathcal{L}_{Dis}$ + $\mathcal{L}_{D}$ &  \textbf{65.7} & \textbf{60.2} & \textbf{52.3} & \textbf{43.4} & \textbf{36.0} & \textbf{36.7}  \\
\bottomrule
\end{tabular}
}
\vspace{-0.2cm}

\end{table}
\begin{table}[t]
\centering
\caption{\label{tab:impact_mi}\textbf{Impact of MI-based denoising} on THUMOS14. Our \proposed{}, employing MI-based pDMI loss in $\mathcal{L}_{D}$ performs favorably compared to utilizing standard losses (\texttt{L1} and \texttt{BCE}) in $\mathcal{L}_{D}$.}
\setlength{\tabcolsep}{14pt}
\adjustbox{width=\columnwidth}{
\begin{tabular}{lccc}
\toprule
  & \texttt{L1} & \texttt{BCE} & \textbf{Ours: \proposed} \\
 \cmidrule(lr){2-4}
 \textbf{mAP} at IoU=$0.5$ & 32.9 & 33.5 & \textbf{36.0} \\
 \bottomrule
\end{tabular}
}
\vspace{-0.2cm}
\end{table}

\subsection{Ablation Study\label{sec:ablation}}
As discussed earlier, our \proposed{} comprises a discriminative $\mathcal{L}_{Dis}$ and a denoising loss $\mathcal{L}_{D}$. Here, we perform comparisons by replacing the two proposed loss terms ($\mathcal{L}_{Dis}$ and $\mathcal{L}_{D}$) in our framework with either the standard cross-entropy loss $\mathcal{L}_{CE}$ or the focal loss $\mathcal{L}_{F}$. In addition, we also show the performance of our \proposed{} with only $\mathcal{L}_{Dis}$. Tab.~\ref{tab:ablation} presents these performance comparisons, in terms of mAP and F1, on THUMOS14. Employing a standard cross-entropy loss ($\mathcal{L}_{CE}$ in Tab.~\ref{tab:ablation}) in our framework results in an mAP score of $23.0$ at IoU=$0.5$. We observe that training with the standard focal loss (obtained by zeroing the weights $w$ in Eq.~\ref{eqn:discriminative_loss}) helps alleviate the issue of a large number of easy samples overwhelming hard samples. This setting, $\mathcal{L}_{F}$ in Tab.~\ref{tab:ablation}, gains $3.7\%$ mAP at IoU=$0.5$ over $\mathcal{L}_{CE}$, thereby  highlighting the need to tackle imbalance between easy backgrounds and hard foregrounds. To the best of our knowledge, we are the first to evaluate the standard focal loss, $\mathcal{L}_{F}$, in weakly-supervised action localization setting.
Our \proposed{} with the discriminative loss term $\mathcal{L}_{Dis}$, which jointly addresses class-imbalance and enhances background-foreground separation, provides consistent improvements over $\mathcal{L}_F$ and achieves $32.2\%$ mAP at IoU=$0.5$. An absolute gain of $5.5\%$ in terms of mAP at IoU=$0.5$ is obtained by the introduction of our proposed $\mathcal{L}_{Dis}$ in place of $\mathcal{L}_F$. Furthermore, our \proposed{} comprising both $\mathcal{L}_{Dis}$ and $\mathcal{L}_{D}$ obtains the best results with an mAP score of $36.0\%$ at IoU=$0.5$. Our  \proposed{} achieves absolute gains of $12.9\%$ and $9.2\%$ in terms of mAP at IoU=$0.5$, over $\mathcal{L}_{CE}$ and $\mathcal{L}_F$, respectively. It is noteworthy that our final \proposed{}, containing both  $\mathcal{L}_{Dis}$ and $\mathcal{L}_{D}$, obtains a significant gain of 5.9\% in terms of F1 score over $\mathcal{L}_{Dis}$ alone. 
This improvement over $\mathcal{L}_{Dis}$ alone is obtained due to explicitly addressing the noise in TCAMs by our $\mathcal{L}_{D}$, leading to a substantial reduction (28\%) in the number of false positives without affecting the recall.

\noindent\textbf{Impact of MI-based denoising:} We also perform an experiment by replacing the proposed pDMI loss in our $\mathcal{L}_{D}$ with the standard \texttt{L1} and \texttt{BCE} losses for denoising the snippet-level activations.  The \texttt{L1} and \texttt{BCE} losses, which do not explicitly capture MI, achieve mAP scores of $32.9\%$ and $33.5\%$ at IoU=$0.5$, respectively, on THUMOS14 (see Tab.~\ref{tab:impact_mi}). Our \proposed{}, which employs MI-based pDMI loss in $\mathcal{L}_{D}$, achieves improved results with an mAP score at IoU=$0.5$ of $36.0\%$. These results suggest that our MI-based denoising is able to robustify the TCAMs in a weakly-supervised setting. 

\noindent\textbf{Qualitative results:}
Fig.~\ref{fig:qual_res_acn} shows a qualitative comparison between the baseline (red) and \proposed{} (blue), along with the ground-truth (GT) action segments (green). 
The baseline employs only $\mathcal{L}_F$ and is the same as the one used in Fig.~\ref{fig:denoise_intuition}. Example test videos with \textit{Diving} and \textit{Throw Discus} actions from THUMOS14 are shown in the first two rows. The baseline incorrectly merges multiple GT instances (\eg, $1$ to $5$ GT in \textit{Diving}) and produces false positives in background regions (\eg, towards the beginning of \textit{Diving} video). Our \proposed{} correctly detects these multiple action instances and suppresses most false positives in the background regions. The third row shows an example test video with \textit{Mowing Lawn} activity from ActivityNet1.2. The baseline incorrectly detects the presence of the activity over the entire video length. In contrast, our \proposed{} improves the detection of multiple activity instances, leading to promising localization performance. %
Additional results and discussions are provided in the appendix.

\section{Conclusion}
We propose a weakly-supervised action localization approach, called \proposed{}, that comprises a discriminative and a denoising loss. 
The discriminative loss term strives for improved foreground-background separability through interlinked classification and localization objectives. The denoising loss term complements the discriminative term by tackling the foreground-background noise in the activations. This is achieved by maximizing the mutual information between activations and labels within a video (intra-video) and across videos (inter-video). Comprehensive experiments performed on multiple benchmarks show that our \proposed{} performs favorably against existing methods on all datasets.

\section*{Acknowledgements}
This work is partially supported by ARC DECRA Fellowship DE200101100, NSF CAREER Grant \#1149783 and VR starting grant 2016-05543.

\bibliographystyle{ieee_fullname}
\bibliography{egbib}

\clearpage

\appendix

Here, we present additional qualitative and quantitative analysis of the weakly-supervised action localization performance of our proposed \proposed{}. The quantitative analysis \wrt robustness and impact of design choices are presented in Sec.~\ref{sec:appn_ablation}, followed by the qualitative results in Sec.~\ref{sec:qual_res}.

\section{Additional Quantitative Analysis\label{sec:appn_ablation}}
In this section, we present additional quantitative results \wrt model sensitivity, ablations and state-of-the-art comparison on the Charades~\cite{charades} dataset.

\noindent\textbf{Ablations for penalty term in $\mathcal{L}_{Dis}$:} 
Here, we present an ablation to analyse the impact of the weights in the penalty term of our proposed discriminative loss term (Eq.~4 in main paper). Tab.~\ref{tab:penalty_ablation} shows the performance comparison on the THUMOS14 dataset for ablating the penalty term. The penalty term in standard focal loss ($\mathcal{L}_F$ in Tab.~\ref{tab:penalty_ablation}) comprises only the prediction dependent term (\eg, $(1-\mathbf{p}[c])$ for a positive class). In contrast, our $\mathcal{L}_{Dis}$ without focal penalty comprises only the grouping and clustering weights (\eg, $(w_{fg}+ w_{fb})$ for a positive class). Furthermore, our final $\mathcal{L}_{Dis}$ includes both the standard focal penalty along with the grouping and clustering weights. Tab.~\ref{tab:penalty_ablation} shows that replacing the standard penalty term with our grouping and clustering weights based penalty term (denoted as $\mathcal{L}_{Dis}$ w/o focal penalty) achieves promising performance over $\mathcal{L}_F$. The performance is further improved in our final $L_{Dis}$, which combines the standard penalty along with our grouping and clustering weights in the penalty term. This shows the efficacy of integrating our grouping and clustering weights ($w_{fg}$, $w_{bg}$ and $w_{fb}$) into the penalty term, for improving the localization.

\begin{table}[t]
\centering
\caption{\label{tab:penalty_ablation}\textbf{Performance comparison} by ablating the penalty term in $\mathcal{L}_{Dis}$, on the THUMOS14 dataset. The penalty term in our $\mathcal{L}_{Dis}$ includes the standard focal loss penalty along with the proposed grouping and separating terms ($w_{fg}$, $w_{bg}$ and $w_{fb}$). In comparison to the standard focal loss $\mathcal{L}_{F}$, our $\mathcal{L}_{Dis}$ without the focal loss penalty term achieves promising performance. This is further improved by our final $\mathcal{L}_{Dis}$, indicating the efficacy of integrating $w_{fg}$, $w_{bg}$ and $w_{fb}$ into the penalty term.
}
\adjustbox{width=\columnwidth}{
\begin{tabular}{lcccccc}
\toprule
\multirow{2}{*}{\textbf{Loss term}} &  \multicolumn{5}{c}{\textbf{mAP @ IoU}}  \\ 
 &  \textbf{0.1} & \textbf{0.2} & \textbf{0.3} & \textbf{0.4} & \textbf{0.5}   \\ \midrule
 $\mathcal{L}_{F}$ &  58.8 & 52.4 & 44.3 & 35.7 & 26.7  \\ 
 $\mathcal{L}_{Dis}$ w/o focal penalty & 62.9 & 57.5 & 47.2 & 37.9 & 29.2  \\
 $\mathcal{L}_{Dis}$ &  65.4 & 59.7 & 50.1 & 40.4 & 32.2   \\
\bottomrule
\end{tabular}
}
\vspace{-0.2cm}

\end{table}

\begin{table}[t]
\centering
\caption{\label{tab:mi_ablation}\textbf{Impact of snippet-level and video-level denoising} on the THUMOS14 dataset. Integrating snippet-level ($\mathcal{L}_{DS}$) and video-level ($\mathcal{L}_{DV}$) denoising terms individually with $\mathcal{L}_{Dis}$ improves the localization performance over $\mathcal{L}_{Dis}$ alone. Moreover, integrating both denoising terms with the discriminative loss term (\ie, $\mathcal{L}_{Dis} + \mathcal{L}_D$) in our \proposed{} achieves improved localization performance, indicating the importance of both snippet-level and video-level denoising for temporal localization.
}
\adjustbox{width=\columnwidth}{
\begin{tabular}{lcccccc}
\toprule
\multirow{2}{*}{\textbf{Loss term}} &  \multicolumn{5}{c}{\textbf{mAP @ IoU}}  \\ 
 &  \textbf{0.1} & \textbf{0.2} & \textbf{0.3} & \textbf{0.4} & \textbf{0.5}   \\ \midrule
  $\mathcal{L}_{Dis}$ &  65.4 & 59.7 & 50.1 & 40.4 & 32.2  \\
 $\mathcal{L}_{Dis} + \mathcal{L}_{DS} $ & 63.0 & 57.1 & 50.1 & 41.9 & 34.3  \\ 
 $\mathcal{L}_{Dis} + \mathcal{L}_{DV}$ & 65.4 & 59.8 & 51.3 & 42.0 & 33.2  \\
 \textbf{\proposed{}} ($\mathcal{L}_{Dis} + \mathcal{L}_{D}$) &  65.8 & 60.1 & 52.3 & 43.4 & 36.0   \\
\bottomrule
\end{tabular}
}
\vspace{-0.2cm}

\end{table}

\noindent\textbf{Impact of snippet-level and video-level denoising:} Tab.~\ref{tab:mi_ablation} shows the impact of 
individually integrating the mutual information (MI) based snippet-level ($\mathcal{L}_{DS}$) and video-level ($\mathcal{L}_{DV}$) denoising terms with $\mathcal{L}_{Dis}$. Integrating both these terms individually improves the localization performance over $\mathcal{L}_{Dis}$ alone. While integrating $\mathcal{L}_{DS}$ achieves $34.3\%$ mAP at IoU${=}0.5$, integrating $\mathcal{L}_{DV}$ suppresses more false positives and results in an mAP of $33.2\%$. Furthermore, our \proposed{}, which integrates both snippet-level and video-level denoising terms with the discriminative loss term (\ie, $\mathcal{L}_{Dis} + \mathcal{L}_D$) achieves improved localization performance, indicating the importance of both snippet-level and video-level denoising for temporal localization.

\begin{table}[t]
\centering
\caption{\label{tab:gamma_ablation}\textbf{Impact of varying $\gamma$} on the THUMOS14 dataset. Sub-optimal localization performances are observed when there is no/very high intra-class grouping, \ie, $\gamma$ is 0 or 1. Promising localization performance is achieved when the intra-class embeddings are coarsely grouped, \ie, $\gamma \in [0.01, 0.1]$.
}
\setlength{\tabcolsep}{12pt}
\adjustbox{width=\columnwidth}{
\begin{tabular}{ccccccc}
\toprule
\multirow{2}{*}{\textbf{Gamma} ($\gamma$)} &  \multicolumn{5}{c}{\textbf{mAP @ IoU}}  \\ 
 &  \textbf{0.1} & \textbf{0.2} & \textbf{0.3} & \textbf{0.4} & \textbf{0.5}   \\ \midrule
  0.0 &  64.8 & 59.3 & 51.8 & 42.5 & 34.2 \\
 0.01 & 65.8 & 60.1 & 52.3 & 43.4 & 36.0  \\ 
 0.1 & 65.5 & 60.0 & 52.0 & 43.1 & 35.7 \\
 1.0 & 65.2 & 59.9 & 51.3 & 41.9 & 33.7   \\
\bottomrule
\end{tabular}
}
\vspace{-0.2cm}

\end{table}

\noindent\textbf{Impact of varying $\gamma$:} Tab.~\ref{tab:gamma_ablation} shows the impact of varying the degree of intra-glass grouping on the THUMOS14 dataset. We observe that when there is no/very high intra-class grouping amongst the foreground embeddings (or background embeddings), the temporal localization of actions is hampered. Furthermore, promising localization performance is achieved when the intra-class grouping is performed at a coarse level, \ie, $\gamma \in [0.01, 0.1]$. This shows that grouping the intra-class embeddings coarsely amongst themselves helps in learning discriminative embeddings, leading to improved localization performance.


\begin{table}[t]
\centering
\caption{\label{tab:sota_charades}\textbf{State-of-the-art comparison} on the Charades dataset. Our \proposed{} performs favorably compared to existing weakly-supervised approaches.}
\adjustbox{width=0.9\columnwidth}{
\begin{tabular}{lccc}
\toprule
  & \texttt{ActGraph}~\cite{rashid2020action} & \texttt{WSGN}~\cite{fernando2020WACV} & \textbf{Ours: \proposed} \\
 \midrule
 \textbf{mAP}  & 15.8 & 18.3 & \textbf{19.2}
\\
 \bottomrule
\end{tabular}
}
\vspace{-0.2cm}
\end{table}

\begin{figure*}[t]
    \centering
    \begin{subfigure}[t]{0.49\textwidth}
        \centering
        \includegraphics[width=\textwidth]{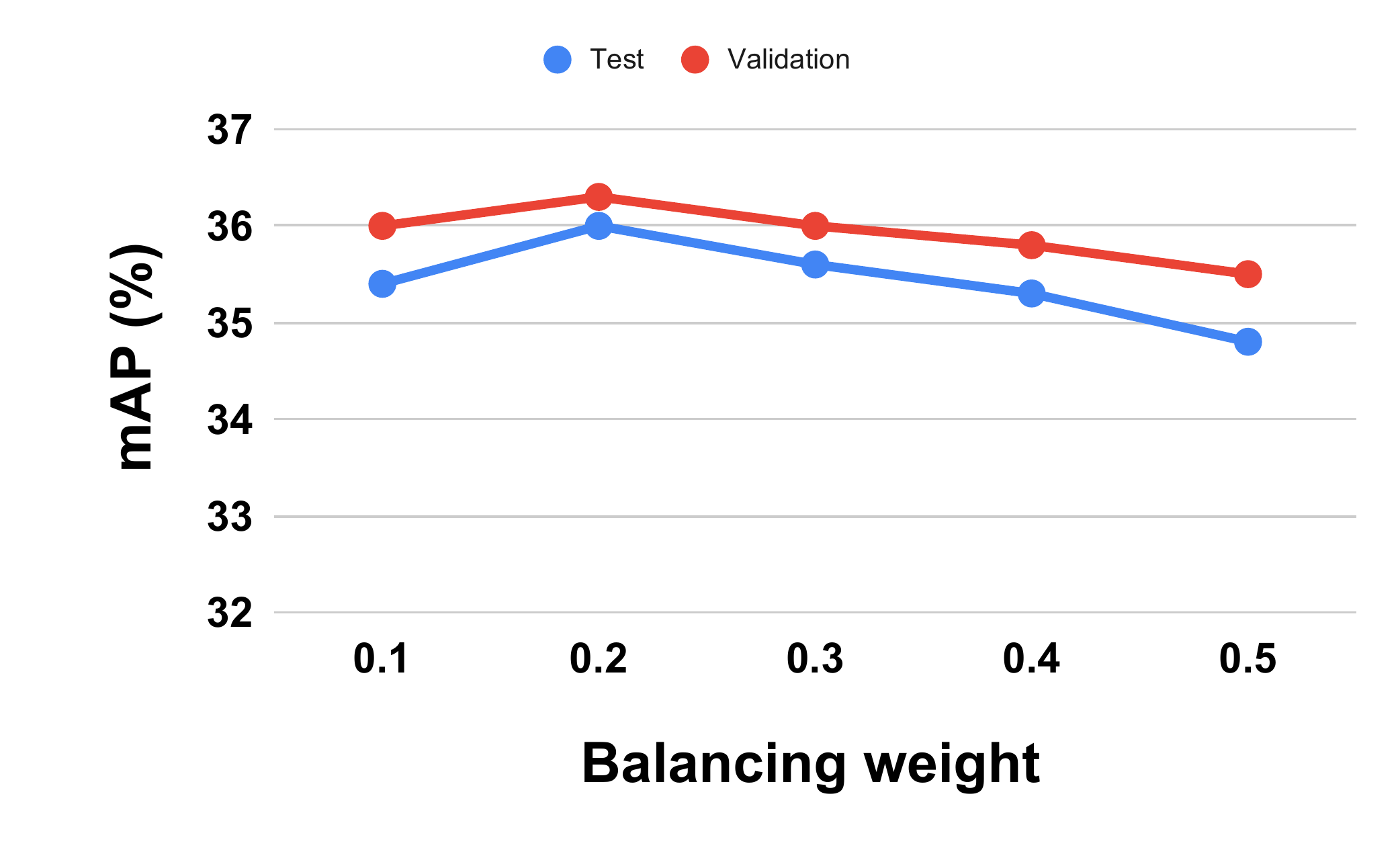}
        \caption{\label{fig:dmi}}
    \end{subfigure}
    \begin{subfigure}[t]{0.49\textwidth}
        \centering
        \includegraphics[width=\textwidth]{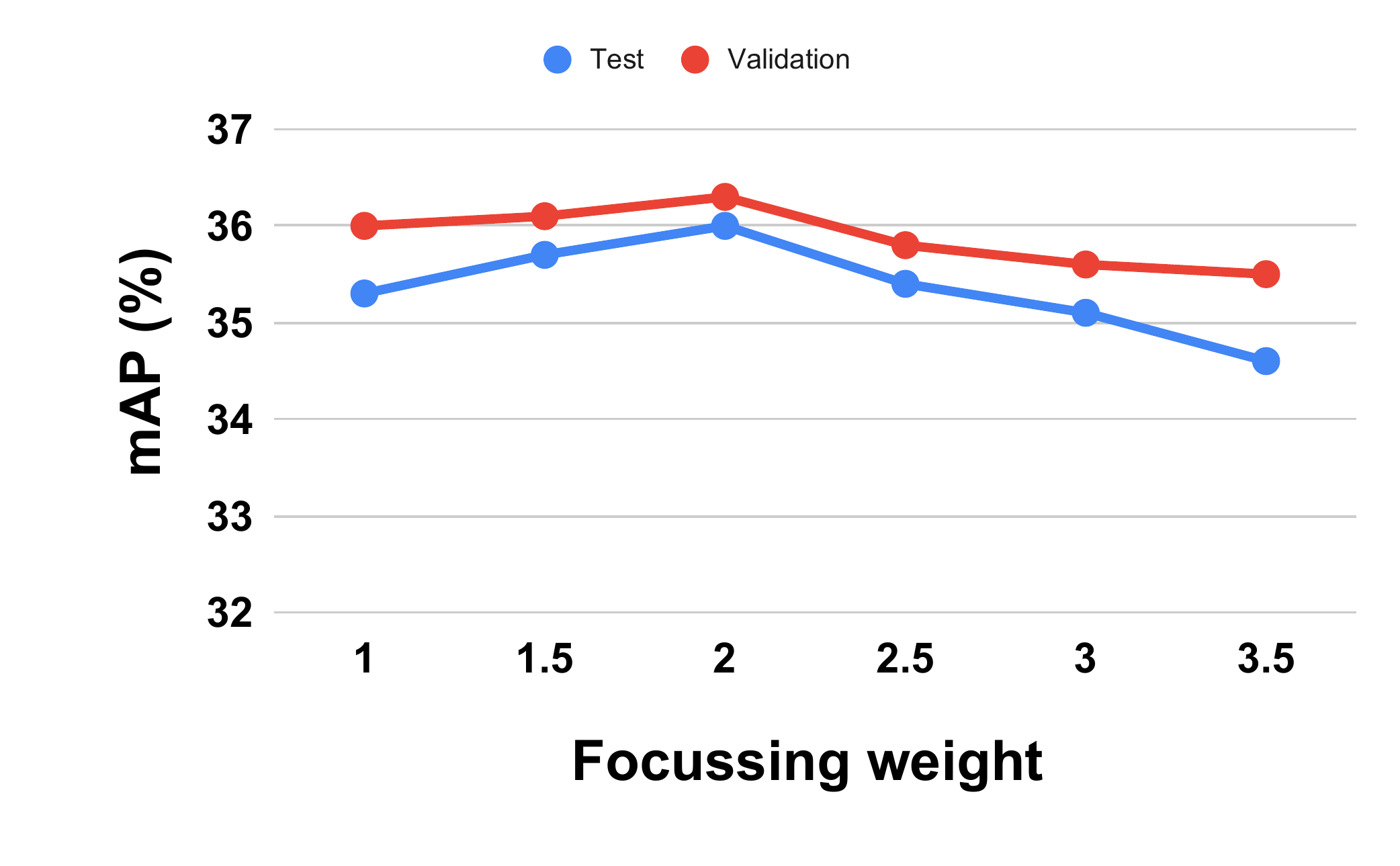}
        \caption{\label{fig:focus}}
    \end{subfigure}%
    \caption{\label{fig:robustness}Action localization performance \wrt balancing parameter $\alpha$ in (a) and focusing parameter $\beta$ in (b) on the THUMOS14 dataset. The performance is shown for both validation and test sets. These experiments show that our \proposed{} is reasonably robust to such variations of the balancing and focusing parameters and achieves promising localization performance.}
\end{figure*}

\noindent\textbf{State-of-the-art Comparison:} The \textbf{Charades}~\cite{charades} dataset comprises $9848$ indoor videos with $157$ everyday activity classes. On an average, there are $6.8$ activity instances per video, with complex activities co-occurring. As in~\cite{async2017cvpr}, we use the standard training and validation split and follow the same localization evaluation. Tab.~\ref{tab:sota_charades} shows the performance comparison of our approach with existing weakly-supervised methods on the Charades dataset. Note that a strongly-supervised approach of \texttt{TGM}~\cite{piergiovanni2019tgm} achieves an mAP of $22.3$. Among the weakly-supervised approaches, the graph convolution networks based \texttt{ActGraph}~\cite{rashid2020action} achieves $15.8\%$ mAP, while Gaussian networks-based \texttt{WSGN}~\cite{fernando2020WACV} obtains $18.3$. Our \proposed{} performs favorably against existing weakly-supervised methods, achieving a promising performance of $19.2$ mAP.

\noindent\textbf{Robustness Analysis:}
Here, we analyse the robustness of our \proposed{} \wrt variations in the balancing parameter $\alpha$ and focusing parameter $\beta$. The performance variations of our approach on both validation and test sets of the THUMOS14 dataset are shown in Fig.~\ref{fig:robustness}. The validation accuracy is obtained through cross-validation. The two parameters $\alpha$ and $\beta$ are varied independently, while keeping the other constant at its respective optimal setting. Varying the balancing weight $\alpha$ results in a performance variation as shown in Fig.~\ref{fig:dmi}. We observe that the performance is optimal when $\alpha$ is around $0.2$ and decreases slowly on either side. As $\alpha$ is increased, the denoising loss term ($\mathcal{L}_D$ in Eq.~1 of main paper) overpowers the discriminative loss ($\mathcal{L}_{Dis}$), resulting in a decreased localization performance. In contrast, as $\alpha$ is decreased, the noise in the temporal class activations remains, resulting in reduced localization performance. Hence, we set $\alpha=0.2$ in our experiments. 
Similarly, an optimal localization performance of $36.0$ mAP is achieved when the focusing parameter $\beta$ is set to $2$ and decreases on either side of it (see Fig.~\ref{fig:focus}). Note that a similar variation in performance is also observed when using the standard focal loss~\cite{focal_loss} for generic object detection. Hence, as in~\cite{focal_loss}, we set $\beta$ as $2$ throughout our experiments. These experiments show that our \proposed{} is reasonably robust to such variations of the balancing and focusing parameters and achieves promising localization performance.

\section{Additional Qualitative Results\label{sec:qual_res}}
\begin{figure*}[t]
    \centering
    \includegraphics[width=\textwidth]{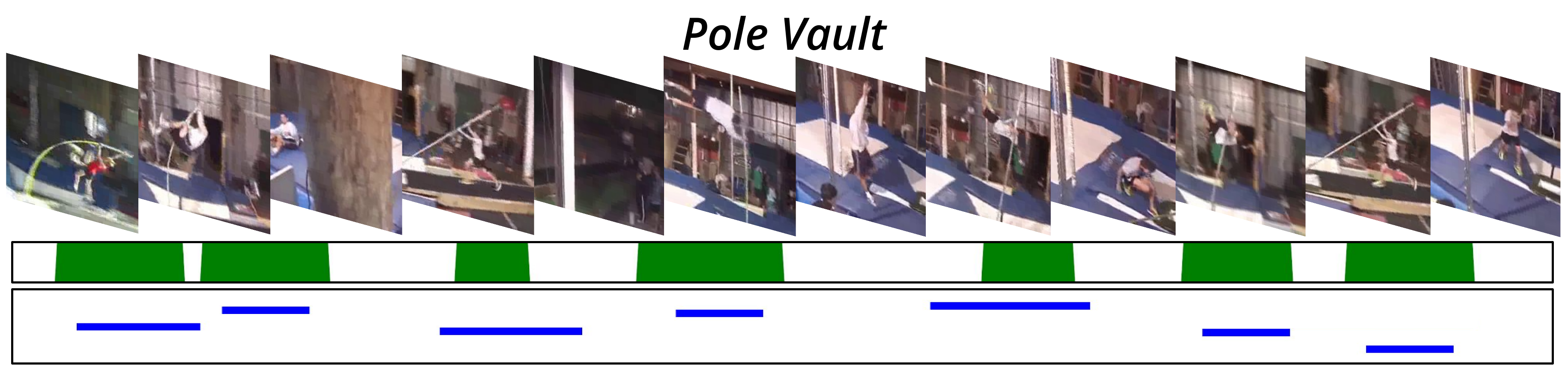}
    \caption{\label{fig:qual_res_thumos1}Well separated action instances of \textit{Pole Vault} are generally accurately detected by our \proposed{}. 
}
\end{figure*}

\begin{figure*}[t]
    \centering
    \includegraphics[width=\textwidth]{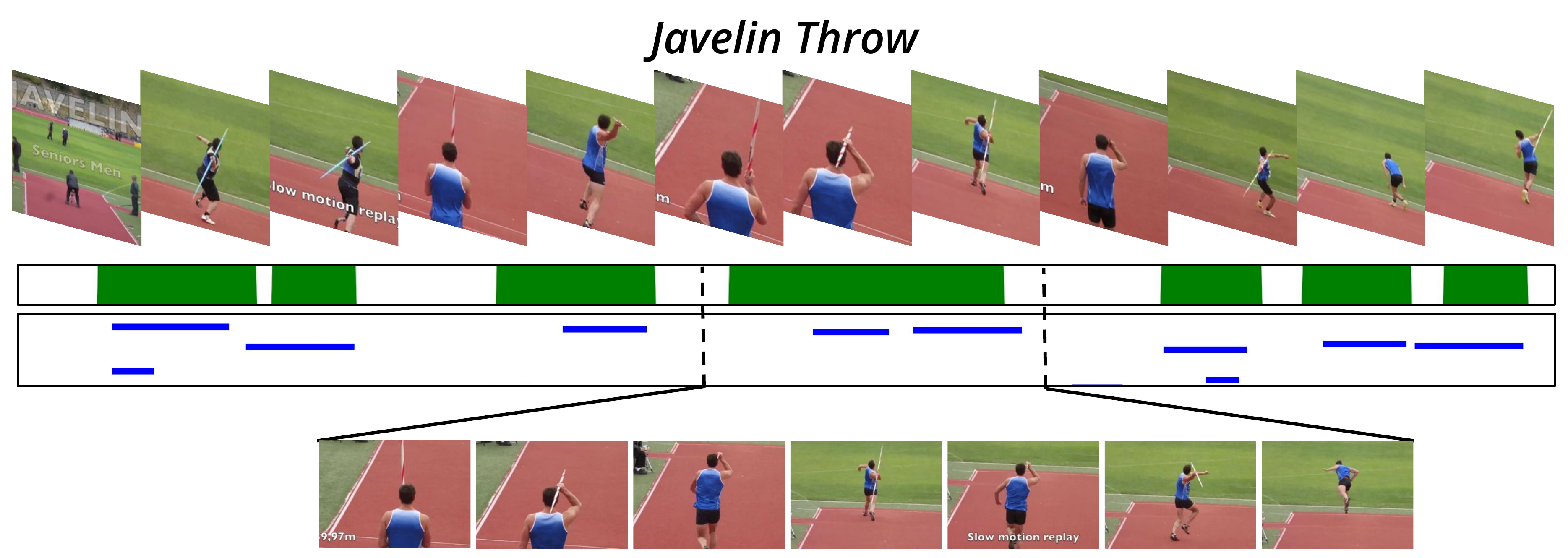}
    \caption{\label{fig:qual_res_thumos2}Fourth instance of \textit{Javelin Throw} is incorrectly annotated as a single instance though it has two instances: action and its slow motion replay. Our \proposed{} correctly detects the two as separate instances. 
 }
\end{figure*}

\begin{figure}[h!]
    \centering
    \includegraphics[width=\columnwidth]{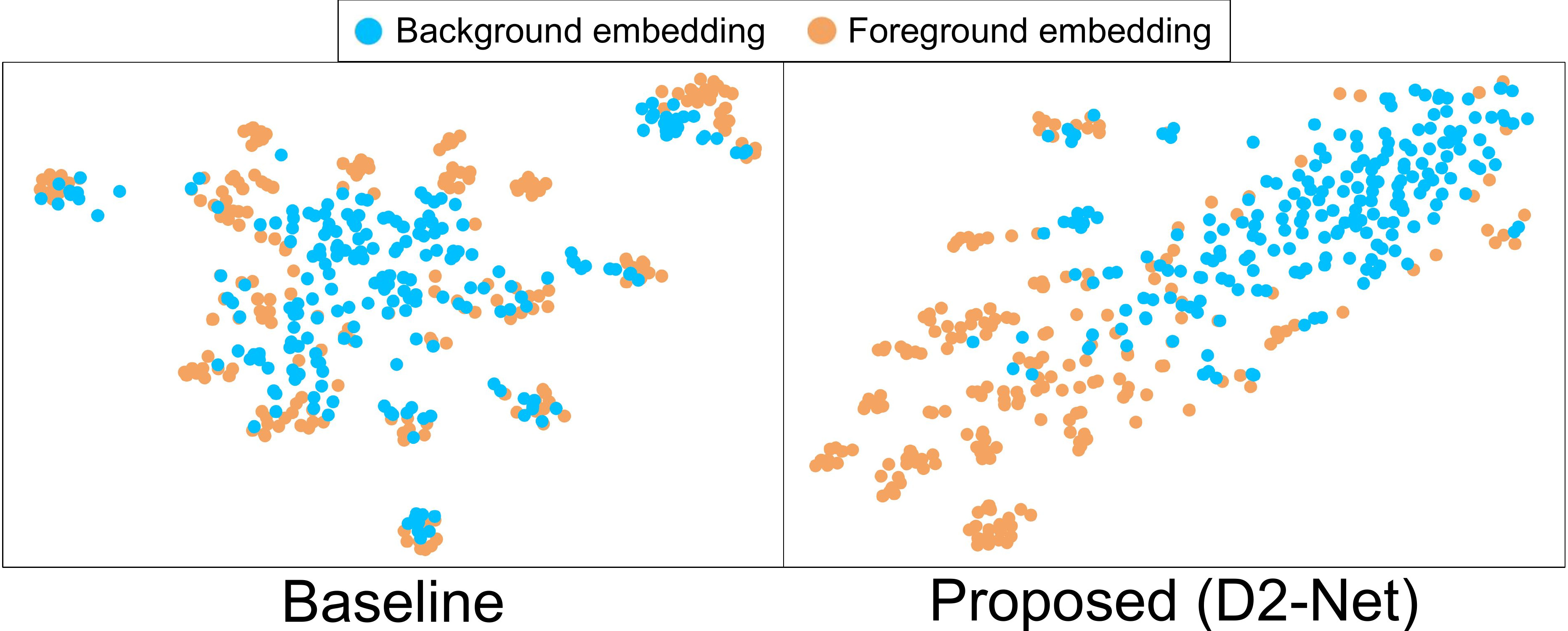}
    \caption{\textbf{Illustration of foreground-background separability} obtained in the latent embedding space of (a) the baseline using the standard focal loss and (b) our \proposed{} via t-SNE scatter plots on the THUMOS14 test set. In both cases, foreground and background embeddings per video are obtained as the mean of latent embeddings at their respective ground-truth locations. Our \proposed{} better separates the foreground and background, compared to the baseline.\vspace{-0.2cm}}
    \label{fig:tsne}
\end{figure}

\begin{figure*}[t]
    \centering
    \includegraphics[width=\textwidth]{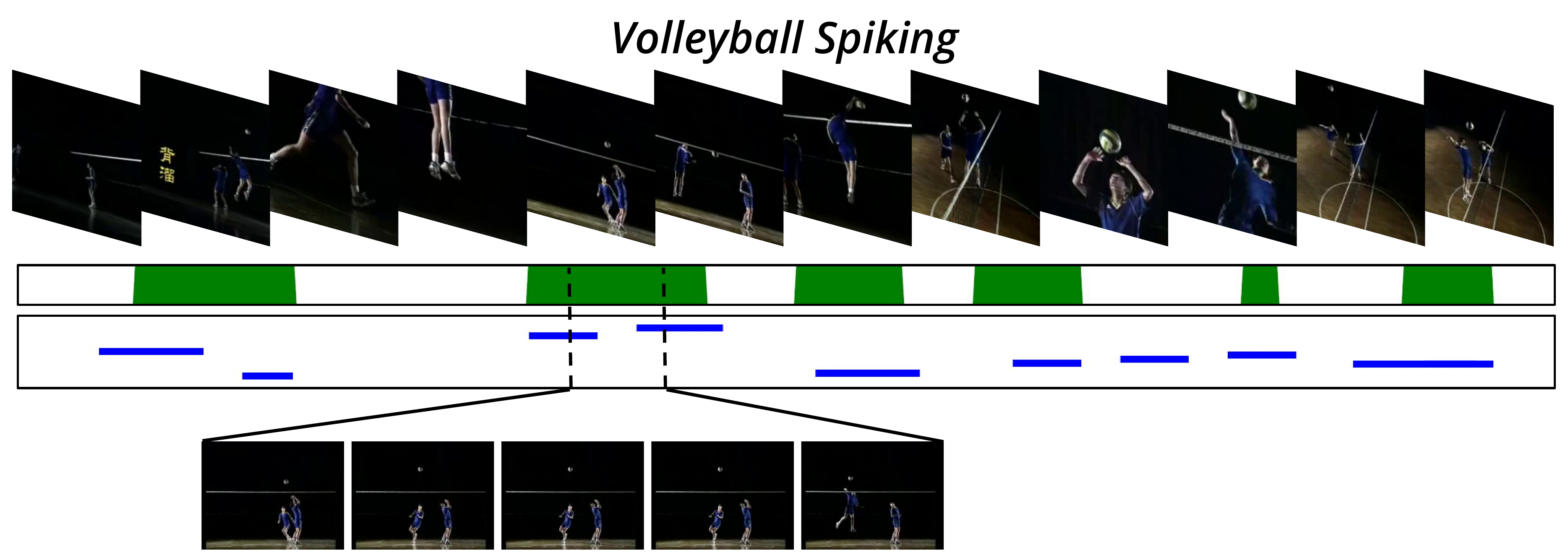}
    \caption{\label{fig:qual_res_thumos4}The first two instances of \textit{Volleyball Spiking} have a considerable pause in the video, resulting in the absence of motion for the corresponding frames. \Eg, an inset of sample frames in the second instance shows the pause in the video containing zero motion. This absence of discriminative motion information leads to four incorrect detections for these two GT instances. 
    }
\end{figure*}

\begin{figure*}[t]
    \centering
    \includegraphics[width=\textwidth]{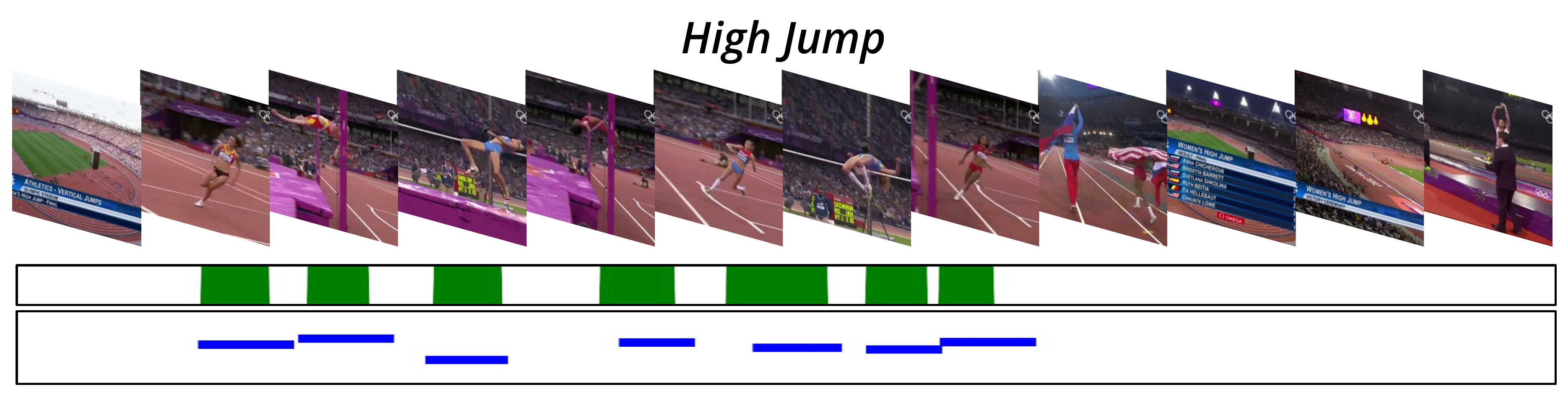}
    \caption{ \label{fig:qual_res_thumos5}Temporally adjacent action instances of \textit{High Jump} (sixth and seventh instances) are correctly detected as distinct instances by our \proposed{}. 
}
\end{figure*}

\begin{figure*}[t]
    \centering
    \includegraphics[width=\textwidth]{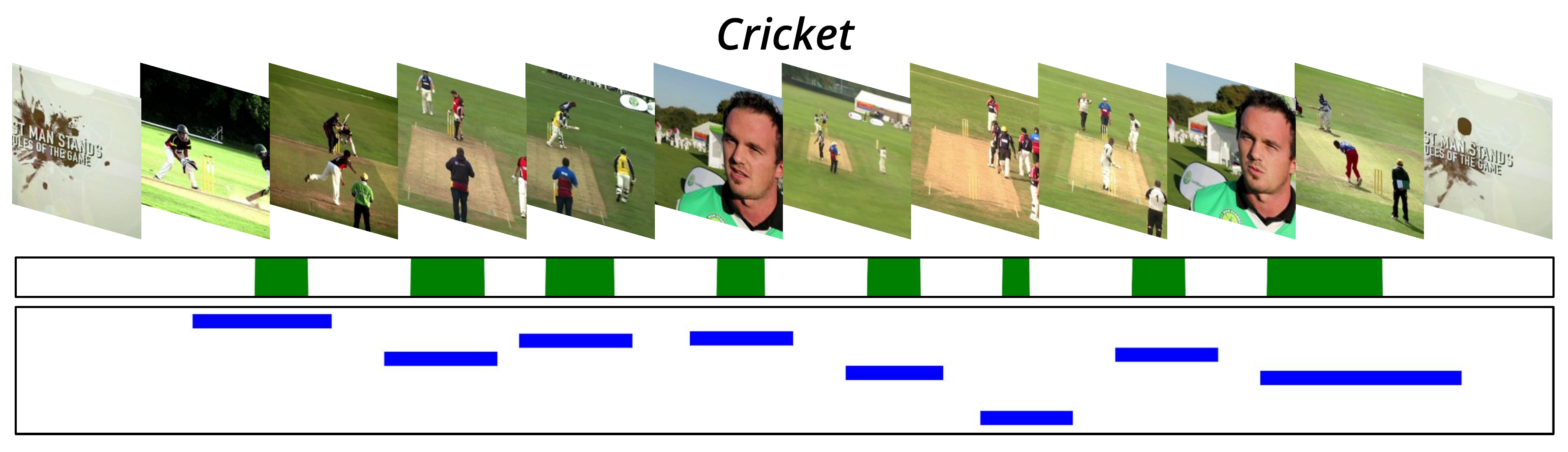}
    \caption{\label{fig:qual_res_acn1}Well separated instances of \textit{Cricket} activity are detected accurately by our \proposed{}.
}
\end{figure*}

\begin{figure*}[t]
    \centering
    \includegraphics[width=\textwidth]{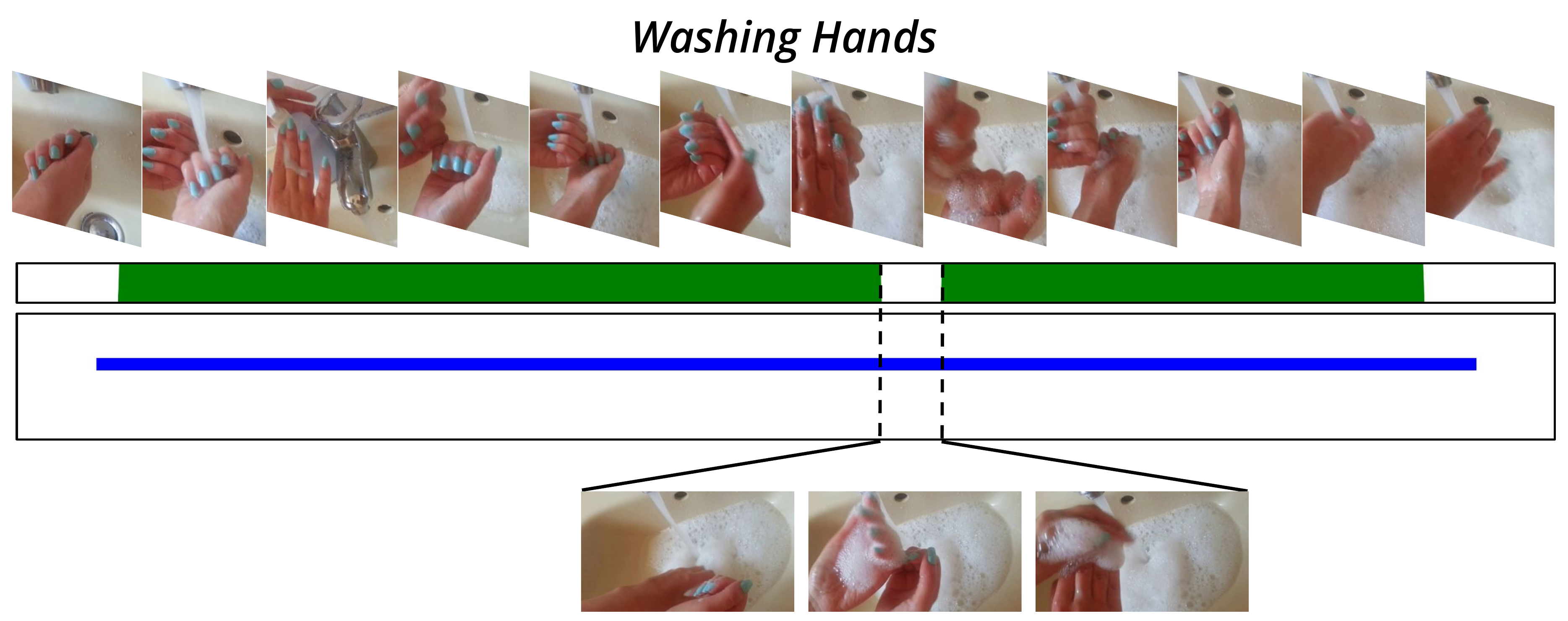}
    \caption{\label{fig:qual_res_acn2}The two adjacent ground-truth \textit{Washing Hands} instances are jointly detected as a single instance by our \proposed{}, since the separating background is indiscriminable from the foreground activity. Sample background frames, shown inset, contain hands along with soap lather and flowing water and are visually similar to the foreground activity.
}
\end{figure*}

\begin{figure*}[t]
    \centering
    \includegraphics[width=\textwidth]{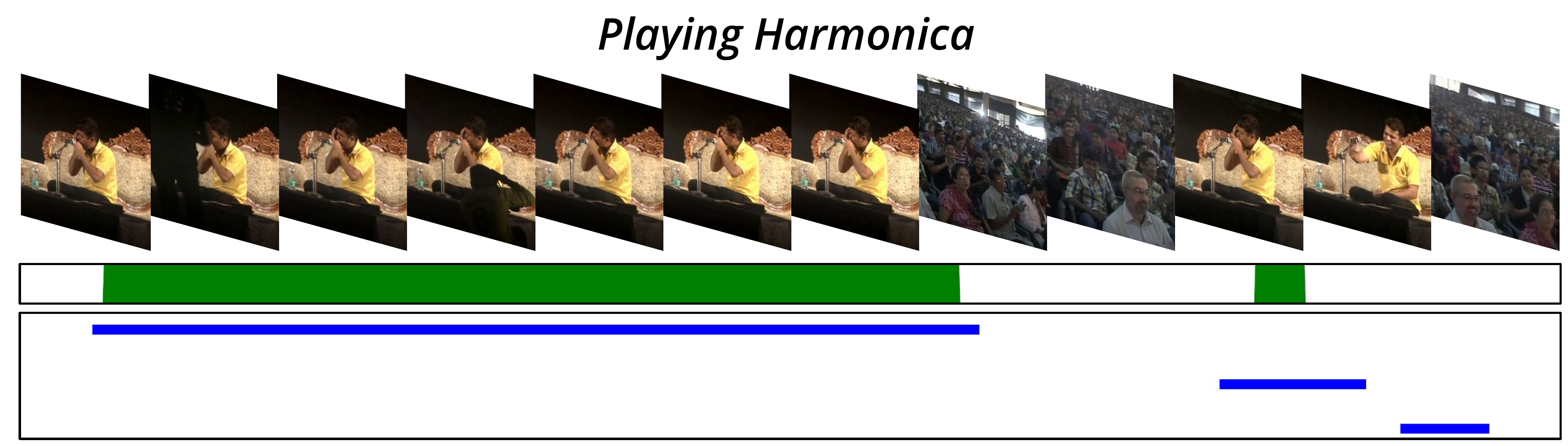}
    \caption{\label{fig:qual_res_acn4}Both the long and short duration instances of \textit{Playing Harmonica} are detected correctly by \proposed{}. However, a false detection arises due to the presence of the performer on stage (but not playing) in the corresponding image frames.
 }
\end{figure*}

\begin{figure*}[t]
    \centering
    \includegraphics[width=\textwidth]{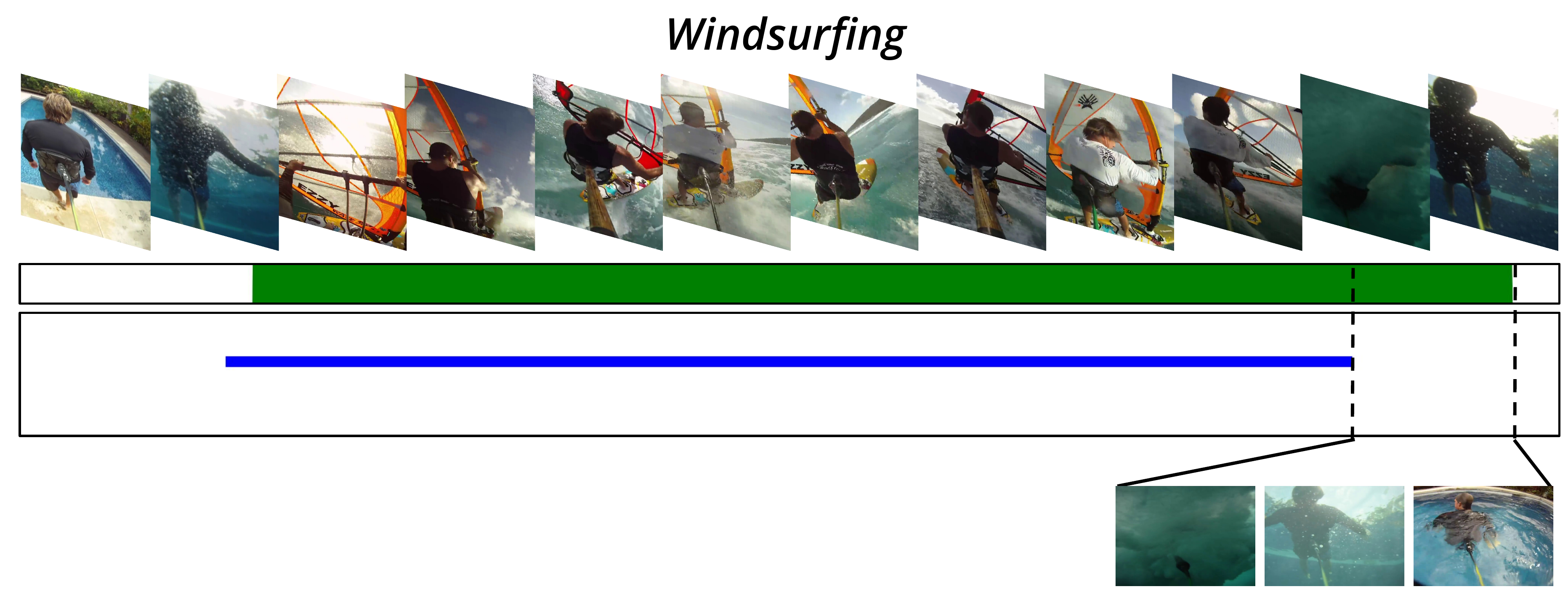}
    \caption{\label{fig:qual_res_acn5}The ground-truth annotation for the end of \textit{Windsurfing} activity is inaccurate since background regions are also included as foreground activity, as shown by the inset frames. Our \proposed{} accurately detects the temporally long activity.
 }
\end{figure*}

Here, we present qualitative temporal action localization results of our \proposed{} framework on example videos from the THUMOS14~\cite{thumos14} and ActivityNet1.2~\cite{activitynet} datasets. In each figure (Fig.~\ref{fig:qual_res_thumos1} to \ref{fig:qual_res_acn5}), sample frames from a video are shown in the top row followed by the ground-truth segments (green) and predicted detections (blue). The height of a detection is indicative of its score. 

\paragraph{THUMOS14:} Fig.~\ref{fig:qual_res_thumos1} to \ref{fig:qual_res_thumos2} and Fig.~\ref{fig:qual_res_thumos4} to \ref{fig:qual_res_thumos5} illustrate the localization results of our \proposed{} on example videos, with \textit{Pole Vault}, \textit{Javelin Throw}, \textit{Volleyball Spiking} and \textit{High Jump} actions from the THUMOS14 dataset. Examples show different scenarios: temporally adjacent instances (\textit{Javelin Throw}, \textit{High Jump}), well separated instances (\textit{Pole Vault}) and action pause (\textit{Volleyball Spiking}). Our \proposed{} detects many of these actions, reasonably well. Generally, well separated actions are detected correctly, as in \textit{Pole Vault} (Fig.~\ref{fig:qual_res_thumos1}). Further, an action instance and its slow motion replay are annotated incorrectly as a single action for the fourth instance in \textit{Javelin Throw} (Fig.~\ref{fig:qual_res_thumos2}), which is correctly detected as two instances by our approach. Accurately detecting the action instances containing video pauses in between, similar to the first and second instances in \textit{Volleyball Spiking} (Fig.~\ref{fig:qual_res_thumos4}), is challenging due to the absence of motion information in the corresponding snippets. The temporally adjacent instances of \textit{High Jump} (Fig.~\ref{fig:qual_res_thumos5}) are correctly delineated. These results show that our approach achieves promising localization performance on these variety of actions.

\paragraph{ActivityNet1.2:} Fig.~\ref{fig:qual_res_acn1} to \ref{fig:qual_res_acn5} illustrate the localization results of our \proposed{} on example videos, with \textit{Cricket}, \textit{Washing Hands}, \textit{Playing Harmonica} and \textit{Windsurfing} actions from the ActivityNet1.2 dataset. Examples show different scenarios: well separated instances (\textit{Cricket}), temporally adjacent activities (\textit{Washing Hands}), long and short activity instances (\textit{Playing Harmonica}), and long activity (\textit{Windsurfing}). Well separated activity instances, similar to the instances of \textit{Cricket} (Fig.~\ref{fig:qual_res_acn1}) are generally detected correctly. The two instances of \textit{Washing Hands} (Fig.~\ref{fig:qual_res_acn2}) are detected as a single instance, since the background that is separating the two instances is indiscriminable from the foreground activity. 
While the long and short activity instances are both detected correctly for \textit{Playing Harmonica} activity (Fig.~\ref{fig:qual_res_acn4}), an additional false detection is observed due to the visual presence of the performer on stage (but not playing) in the corresponding image frames. Though the annotation for the end of \textit{Windsurfing} activity is inaccurate and includes background regions also as foreground activity, our \proposed{} correctly detects the end of the temporally long activity (Fig.~\ref{fig:qual_res_acn5}). These qualitative results show that our proposed approach achieves promising action localization performance on a variety of activities.


\noindent\textbf{Foreground-Background Separation:}
Fig.~\ref{fig:tsne} shows the foreground-background separability comparison, utilizing t-SNE scatter plots, between the baseline and our \proposed{}. Here, foreground and background embeddings per video are obtained by average pooling (temporally) the latent embeddings at their respective ground-truth snippet locations. Fig.~\ref{fig:tsne} shows that the foreground and background embeddings in the baseline overlap with each other. In contrast, our \proposed{} better separates the foreground and background, compared to the baseline, leading to improved localization of foreground actions in the videos.  

\end{document}